\documentclass[preprint,3p,11pt,authoryear]{elsarticle}

\usepackage{setspace}
\usepackage{newtxtext,newtxmath}

\usepackage{microtype}

\usepackage[dvipsnames]{xcolor}
\usepackage{graphicx}
\usepackage{amsfonts}
\usepackage{hyperref}
\usepackage{cleveref}
\usepackage{enumitem}
\usepackage{algpseudocode}
\usepackage{algorithm}
\usepackage{subfig}
\usepackage{caption}
\usepackage{pgfplots}
\usepackage{siunitx}

\usepackage{float}

\algtext*{EndIf}
\algtext*{EndFor}
\algtext*{EndWhile}
\algtext*{EndFunction}
\algtext*{EndProcedure}

\usepackage{tikz}
\usepackage{tikz-3dplot}
\usetikzlibrary{quotes,angles,shapes.misc}
\tikzset{
    cross/.pic = {
    \draw[rotate = 45] (-#1,0) -- (#1,0);
    \draw[rotate = 45] (0,-#1) -- (0, #1);
    }
}

\usepackage{booktabs}
\usepackage{multicol}
\usepackage{enumitem}
\usepackage{xcolor}

\usepackage[utf8]{inputenc}
\usepackage[english]{babel}
\usepackage[T1]{fontenc}

\pgfplotsset{compat=1.18}

\newlength\figurewidth
\newlength\figureheight
\usetikzlibrary{positioning,shapes.multipart,hobby,calc,arrows.meta}

\tikzset{>=stealth,parent node/.style={rectangle split, rectangle split parts=2,align=left,text width=2.5cm,draw,node distance=1cm and 1cm}}

\DeclareMathOperator*{\argmin}{argmin}

\newcommand{\algrule}[1][.1pt]{\par\vskip.3\baselineskip\hrule height #1\par\vskip.3\baselineskip}

\newcommand{\Cmax}{\ensuremath{\text{T}_{\max}}}
\newcommand{\Vmax}{\ensuremath{\text{V}_{\max}}}
\newcommand{\Amax}{\ensuremath{\text{A}_{\max}}}
\newcommand{\Fmax}{\ensuremath{\text{F}_{\max}}}

\newcommand{\norm}[1]{\left\lVert#1\right\rVert}
\newcommand{\card}[1]{\lvert#1\rvert}
\newcommand{\tdots}{%
  \mathinner{{\ldotp}{\ldotp}}%
}
\newcommand{\nsub}{\ensuremath{n_{\text{sub}}}}

\usetikzlibrary{positioning, arrows.meta, shapes,fit,calc}

\tikzset{
    block/.style={
        rectangle, 
        draw=black, 
        fill=white, 
        thick, 
        rounded corners, 
        minimum width=2cm, 
        minimum height=0.5cm, 
        align=center,
        font=\small
    },
    container/.style={
        rectangle,
        draw=black,
        thick,
        rounded corners,
        inner sep=5mm,
        label={[anchor=north west, xshift=0mm, yshift=0mm, font=\bfseries\footnotesize]north west:#1}
    },
    arrow/.style={
        thick,
        -{Latex[length=3mm]}
    }
}

\usepackage{scalerel}
\usetikzlibrary{svg.path}
\definecolor{orcidlogocol}{HTML}{A6CE39}
\tikzset{
orcidlogo/.pic={
  \fill[orcidlogocol] svg{M256,128c0,70.7-57.3,128-128,128C57.3,256,0,198.7,0,128C0,57.3,57.3,0,128,0C198.7,0,256,57.3,256,128z};
  \fill[white] svg{M86.3,186.2H70.9V79.1h15.4v48.4V186.2z}
               svg{M108.9,79.1h41.6c39.6,0,57,28.3,57,53.6c0,27.5-21.5,53.6-56.8,53.6h-41.8V79.1z M124.3,172.4h24.5c34.9,0,42.9-26.5,42.9-39.7c0-21.5-13.7-39.7-43.7-39.7h-23.7V172.4z}
               svg{M88.7,56.8c0,5.5-4.5,10.1-10.1,10.1c-5.6,0-10.1-4.6-10.1-10.1c0-5.6,4.5-10.1,10.1-10.1C84.2,46.7,88.7,51.3,88.7,56.8z};
}
}
\newcommand\orcidicon[1]{\href{https://orcid.org/#1}{\mbox{\scalerel*{
\begin{tikzpicture}[yscale=-1,transform shape]
\pic{orcidlogo};
\end{tikzpicture}
}{|}}}}

\Crefname{section}{Section}{Sections}
\Crefname{figure}{Figure}{Figures}
\Crefname{algorithm}{Algorithm}{Algorithms}
\Crefname{table}{Table}{Tables}
\crefname{section}{Sec.}{Sec.}
\crefname{figure}{Fig.}{Fig.}
\crefname{algorithm}{Alg.}{Alg.}
\crefname{table}{Tab.}{Tab.}

\begin{document}
\begin{frontmatter}
\title{Dynamical Vehicle Orienteering Problem for Multi-Rotor Unmanned Aerial Vehicles}
\author{František Nekovář\corref{cor1}}
\ead{nekovfra@fel.cvut.cz}
\author{Matej Novosad\corref{}}
\ead{novosma2@fel.cvut.cz}
\author{Martin Saska\corref{}}
\ead{saskama1@fel.cvut.cz}
\author{Robert Pěnička\corref{}}
\ead{penicrob@fel.cvut.cz}
\cortext[cor1]{Corresponding author}

\address{Faculty of Electrical Engineering, Czech Technical University, Technicka 2, 166 27, Prague, Czech Republic}

\journal{European Journal of Operational Research}

\date{\today}

\begin{abstract}
This paper introduces the Dynamical Vehicle Orienteering Problem (DVOP), a generalization of the Orienteering Problem (OP).
The OP maximizes the reward collected from spatial targets under a limited travel budget; the DVOP extends it by accounting for both external and vehicle-actuated forces.
We study the DVOP in the context of multi-rotor Unmanned Aerial Vehicle (UAV) flight planning, using a three-dimensional Point-Mass Model (PMM) constrained by maximum velocity and acceleration magnitudes and subject to gravitational acceleration, with the travel budget expressed as a maximum flight time.
Because the DVOP couples reward maximization with time-optimal trajectory planning, it cannot be formulated as a simple graph problem and solved exactly without relaxing or under-actuating the vehicle dynamics.
We therefore propose two solution approaches: a Branch-and-Bound (BnB) procedure that combines Non-Linear Programming (NLP) and Mixed-Integer Linear Programming (MILP) to provide high-quality solutions, and a Large Neighborhood Search (LNS) metaheuristic that supplies an initial reward bound and scales to instances intractable for the BnB.
The BnB relies on a novel MILP formulation of travel costs based on minimum-time trajectory primitives through target triplets, yielding a tight reward upper bound, while the LNS uses limited thrust decomposition to compute fast, high-quality PMM trajectories.
Experiments on benchmark instances show improvements of up to \SI{37}{\percent} over state-of-the-art solutions for the Kinematic Orienteering Problem, and a real-world deployment on a multi-rotor UAV verifies the proposed PMM solution trajectories.
\end{abstract}

\end{frontmatter}


\section{Introduction}

\begin{figure}[ht]
    \centering
    \includegraphics[width=0.5\linewidth]{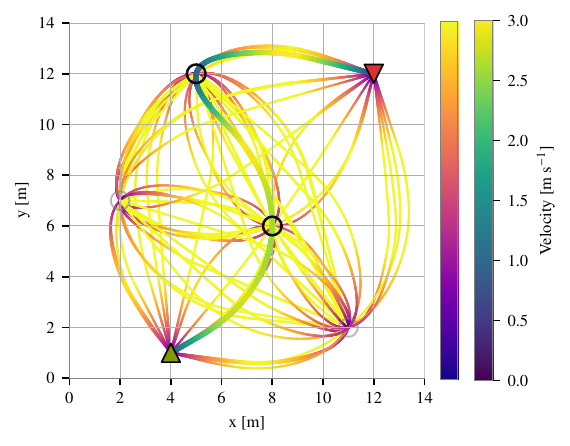}
    \caption{Visualization of individual PMM trajectory primitives used to precompute the MILP cost coefficients for reward upper-bounds during the BnB solution procedure. The green and red triangle markers indicate the start and end positions, and the black and gray circles indicate visited and unvisited targets.
    Velocities of the time-optimal trajectory primitives and the time-optimal solution trajectory through the visited targets are illustrated by the colorbars.}
    \label{fig:titlepage}
\vspace{-1em}
\end{figure}

This paper introduces the \emph{Dynamical Vehicle Orienteering Problem}~(DVOP), which generalizes the \emph{Kinematic Orienteering Problem}~(KOP)~\citep{meyer2022kinematic} and, consequently, the classical \emph{Orienteering Problem}~(OP)~\citep{golden1987orienteering} by explicitly accounting for second-order vehicle dynamics.
In particular, the DVOP considers both the control forces generated by the vehicle and the external forces acting upon it, which is important for multi-rotor aerial vehicles whose motion is actuated by propeller thrust and strongly affected by gravity.

The OP belongs to the family of \emph{Vehicle Routing Problems}~(VRPs)~\citep{mor2022vehicle}.
Its objective is to maximize the reward collected from a set of targets under a maximum travel budget.
In the standard OP, the route is planned on a graph whose nodes represent targets and whose edge costs are typically Euclidean distances, corresponding to straight-line motion with the travel cost equal to the sum of selected edge costs.
This representation is inadequate for physical vehicles, as it requires either fully stopping at each target or infinite acceleration to change direction.
The DVOP addresses this limitation with an acceleration-limited, second-order motion model referred to as the \emph{Point-Mass Model}~(PMM) in the following text.
This makes the DVOP challenging, as it couples two NP-hard problems: the OP and time-optimal PMM trajectory planning.

In contrast to the \emph{Dubins Orienteering Problem}~(DOP)~\citep{pvenivcka2017dubins}, which uses the Dubins vehicle model suited to fixed-wing aircraft, the DVOP employs the PMM to capture the dynamics of a multi-rotor \emph{Unmanned Aerial Vehicle}~(UAV).
The PMM was introduced to the OP in the 2D KOP~\citep{meyer2022kinematic}, where axis-based constraints were imposed on the maximum acceleration and velocity, and the velocity vector magnitude and direction were discretely sampled at each target, under-actuating the available dynamics.
A later model-based KOP formulation~\citep{nekovavr2023multi} imposed velocity and acceleration magnitude constraints and was solved using \emph{Non-Linear Programming}~(NLP); however, discrete time-step sampling again under-actuated the dynamics, forcing the vehicle to slow down at target visits.
The variable time-step extension~\citep{ghotavadekar2024variable} partially alleviated this, but both model-based approaches only approximate target visits, and solving the resulting non-convex NLP does not guarantee global optimality.
Moreover, none of the existing approaches explicitly models the vehicle thrust and gravity as the main forces acting on the vehicle.

To address these limitations, our contributions are as follows.
We formulate the DVOP for a PMM vehicle under gravity for multi-rotor UAV mission planning.
We propose an exact \emph{Branch-and-Bound} (BnB) procedure that combines NLP-based time-optimal trajectory generation with a novel \emph{Mixed-Integer Linear Programming} (MILP) formulation to compute reward upper bounds.
The MILP is based on pre-computed PMM trajectory primitives (\cref{fig:titlepage}) and, as we show experimentally, provides a mean reward upper-bound gap under \SI{8}{\percent} for all valid solution-tree nodes in all evaluated scenarios.
We propose a \emph{Large Neighborhood Search} (LNS) metaheuristic that performs continuous velocity optimization at targets based on \emph{Limited Thrust Decomposition} (LTD)~\citep{teissing2024real}.
The LNS solves instances intractable for the BnB and also provides a warm start (initial reward bound) for it.
We evaluate the solvers on KOP benchmarks and report improvements over the state of the art~\citep{meyer2022kinematic,ghotavadekar2024variable} of up to~\SI{37}{\percent}.
Additional experiments study performance scaling with respect to instance parameters (e.g., the travel budget and acceleration constraint) and establish a baseline for future benchmarking.
We show that the LNS--BnB solution-quality gap remains below \SI{20}{\percent} in all benchmarked instances and decreases as the travel budget grows.
Finally, a real-world deployment confirms the suitability of the solvers for multi-rotor UAV mission planning.

The paper is organized as follows.
\Cref{sec:related} reviews related work; \cref{sec:statement} defines the DVOP; \cref{sec:vns} describes the LNS metaheuristic and \cref{sec:se} the BnB procedure; \cref{sec:results} presents the results; and \cref{sec:conclusion} concludes.

\section{Related works}\label{sec:related}

The DVOP generalizes the KOP~\citep{meyer2022kinematic} and the OP~\citep{golden1987orienteering}, which themselves generalize the Traveling Salesman Problem~\citep{flood1956traveling} and the 0-1 Knapsack Problem~\citep{kolesar1967branch}.
For a broader treatment of the OP and its variants, we refer the reader to the surveys~\citep{VANSTEENWEGEN20111,gunawan2016orienteering}; UAV-related works are discussed below.

Recent OP generalizations unrelated to the DVOP include reconnaissance mission planning with time-varying rewards~\citep{dasdemir2022uav}, a variant of the Dynamic OP~\citep{psaraftis1988dynamic} in which the problem parameters rather than the vehicle dynamics change over time.
Other examples are the \emph{Team Orienteering Problem} (TOP) with UAVs~\citep{peyman2024sim}, which models stochastic travel times from weather conditions, and the truck-assisted OP with UAVs~\citep{morandi2024orienteering}, another TOP variant formulated as a MILP and solved by branch-and-cut~\citep{bnb_original_paper}.

Much of the recent OP literature on motion-constrained aerial routing uses the Dubins vehicle model~\citep{dubins1957dubins}, which assumes constant velocity with a bounded path curvature.
The DOP was first formulated by~\cite{pvenivcka2017dubins} by discretizing the vehicle heading at visited targets, followed by a self-organizing-map formulation~\citep{faigl2017dubinsSOM,kohonen2001tsp_som}.
The discretized-solution-space approach was generalized to the Set Orienteering Problem~\citep{pvenivcka2019variable} and solved with a \emph{Variable Neighborhood Search}~(VNS)~\citep{mladenovic1997variable}, which a variable-speed DOP extension~\citep{faria2024vsdop} later adopted, further expanding the decision space by discretizing the velocity at visited targets.
In the Close Enough DOP, where the vehicle only passes within a prescribed vicinity of the target, VNS remains popular~\citep{penicka2017dopn,faigl2017dceop}, alongside self-organizing-map approaches~\citep{faigl2016dopnsom}.
The Dubins TOP~\citep{vana2017dapop}, solved with VNS, generalizes the DOP to multiple aerial vehicles and was later solved with a Greedy Randomized Adaptive Search Procedure metaheuristic~\citep{zahradka2019dtopn,Souriau2008AGR} and an exact branch-and-price solver~\citep{sundar2022branch,boussier2007exact}.
While suited to fixed-wing aircraft, the purely kinematic Dubins model cannot describe multi-rotor UAV dynamics and does not capture vertical motion, precluding 3D applicability.
To address maneuverability, \cite{faigl2019bezierop} parametrized UAV trajectories with B\'ezier curves under velocity and acceleration constraints and solved them with a self-organizing-map solver, but B\'ezier smoothing causes trajectories to miss exact target locations, restricting the approach to Close Enough formulations.

To overcome the DOP's limitations, \cite{meyer2022kinematic} formulated the KOP, discretizing both heading and velocity magnitude at each target while using second-order \emph{Point-Mass Model} (PMM) dynamics to derive time-optimal flight times between velocity samples, and solved it with an LNS solver~\citep{shaw1998using}.
A key advantage of the PMM is that time-optimal trajectories between two velocity samples are analytic, but only in the single-axis case.
Extension to full 3D motion requires synchronizing per-axis trajectories, also analytic by scaling down the acceleration of faster axes.
For agile UAVs, however, equal per-axis constraints underutilize the available collective thrust.
To address this, \cite{Meyer2023topuav} evaluated trajectories under four predefined constraint distributions, reducing duration by up to 15\%, and \cite{teissing2024real} introduced the \emph{Limited Thrust Decomposition} (LTD) algorithm, which iteratively redistributes constraints until the collective thrust is fully utilized, cutting duration by a further 20\% over~\cite{Meyer2023topuav}.
The same work also addressed multi-waypoint PMM trajectory generation via gradient-based optimization of via-waypoint velocities, providing near-optimal solutions in real time, though without global optimality guarantees.
Subsequently, \cite{nekovavr2023multi} reformulated the KOP with velocity and acceleration magnitude constraints within an NLP framework, and \cite{ghotavadekar2024variable} generalized this formulation with a variable time-step and extended it to 3D.

\section{Problem statement}\label{sec:statement}

The DVOP is motivated by applying the OP to quad-rotor aerial vehicles.
The vehicle model is presented in~\cref{sec:statement:model}, and the DVOP is formulated in~\cref{sec:statement:math}.

\subsection{Point-mass vehicle model with gravity acceleration}
\label{sec:statement:model}
\tdplotsetmaincoords{70}{120}
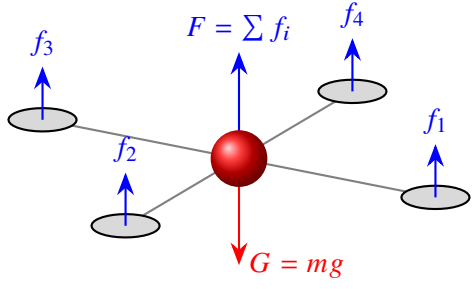
\begin{figure}[h]
    \centering
    \begin{tikzpicture}[
        scale=1.5,
        axis/.style={->,thick},
        force/.style={-{Stealth[length=3mm]},thick,blue},
        gravity/.style={-{Stealth[length=3mm]},thick,red}]
        
        \def\armlen{2}
        \def\forcelen{0.5}

        \begin{scope}[tdplot_main_coords]
        
        \draw[thick,gray] (-\armlen,0,0) -- (\armlen,0,0);
        \draw[thick,gray] (0,-\armlen,0) -- (0,\armlen,0);
        
        \foreach \x/\y in {-\armlen/0, \armlen/0, 0/-\armlen, 0/\armlen} {
            \draw[thick,fill=gray!30] (\x,\y,0) circle (0.3);
        }
        
        \draw[force] (0,\armlen,0) -- (0,\armlen,\forcelen);
        \node[blue,above] at (0,\armlen,\forcelen) {$f_1$};
        
        \draw[force] (\armlen,0,0) -- (\armlen,0,\forcelen);
        \node[blue,above] at (\armlen,0,\forcelen) {$f_2$};
        
        \draw[force] (0,-\armlen,0) -- (0,-\armlen,\forcelen);
        \node[blue,above] at (0,-\armlen,\forcelen) {$f_3$};
        
        \draw[force] (-\armlen,0,0) -- (-\armlen,0,\forcelen);
        \node[blue,above] at (-\armlen,0,\forcelen) {$f_4$};

        \draw[force,thick] (0,0,0) -- (0,0,1.0);
        \node[blue,above] at (0,0,1.0) {$F = \sum f_i$};
        
        \draw[gravity] (0,0,0) -- (0,0,-1.0);
        \node[red,right] at (0,0,-1.0) {$G=mg$};

        \end{scope}

        \shade[ball color=red!100] (0,0,0) circle (0.25);
        
        
    \end{tikzpicture}
    \caption{Illustration of forces acting upon the quad-rotor vehicle of mass $m$, where $f_1 \tdots f_4$ are the individual thrust forces produced by the propeller with the sum denoted as $F$ and $G$ is the force of gravity.}
    \label{fig:uav_forces}
    \vspace{-1em}
\end{figure}
We use the point-mass model of~\cite{teissing2024real} to capture the dynamics of the quad-rotor UAV, actuated by the collective motor thrust $F$ under the gravitational force $G$, as illustrated in~\cref{fig:uav_forces}.
The \emph{Ordinary Differential Equations}~(ODEs) of the vehicle system are described by~\cref{eq:model:ode} as a function of time $t$:
\begin{subequations}\label{eq:model:ode}
\begin{align}
    \dot{\mathbf{p}}(t) &= \mathbf{v}(t),\\
    \dot{\mathbf{v}}(t) &= \mathbf{a}(t).
\end{align}
\end{subequations}
The system states are the vehicle position $\mathbf{p}(t)\in\mathbb{R}^3$ and velocity $\mathbf{v}(t)\in\mathbb{R}^3$, and the input is the acceleration $\mathbf{a}(t)=\dot{\mathbf{v}}(t)\in\mathbb{R}^3$.
The corresponding motor thrust acceleration is $\mathbf{a}(t)-\mathbf{g}$.
\begin{subequations}\label{eq:constraints}
    \begin{align}
        \norm{\mathbf{v}} &\leq \Vmax\label{eq:velmag}\\
        \norm{\mathbf{a}-\mathbf{g}} &\leq \Amax\label{eq:accmag}
    \end{align}
\end{subequations}
\Cref{eq:velmag} constrains the velocity magnitude by $\Vmax\in\mathbb{R}_{>0}$, and~\cref{eq:accmag} constrains the acceleration magnitude by $\Amax\in\mathbb{R}_{>0}$.
Given the vehicle mass $m$, the maximal propeller thrust force $\Fmax$, and the gravitational acceleration $\mathbf{g} = [0,0,-g]^{\top}$, the maximal thrust acceleration is $\Amax = \Fmax \div m$, so~\cref{eq:accmag} limits the magnitude of the thrust acceleration $\mathbf{a}(t)-\mathbf{g}$.

Since the continuous ODE model~\eqref{eq:model:ode} is linear time-invariant, it can be exactly discretized as \emph{Algebraic Difference Equations}~(ADEs) in~\cref{eq:model:ade}:


\begin{subequations}
\label{eq:model:ade}
\begin{align}
    \mathbf{p}_{k+1} &= \mathbf{p}_k + \mathbf{v}_{k} \Delta t + \frac{1}{2}\mathbf{a}_k \Delta t^2,\\
    \mathbf{v}_{k+1} &= \mathbf{v}_k + \mathbf{a}_{k} \Delta t,
\end{align}
\end{subequations}

where $\Delta t \in \mathbb{R}_{>0}$ is the sampling time-step and $k$ the step index.
The constraints in~\cref{eq:velmag,eq:accmag} apply to the discretely sampled values.
The discrete model~\eqref{eq:model:ade} underlies the trajectory optimizations in~\cref{sec:vns,sec:se}.

\subsection{Formulation of the Dynamical Vehicle Orienteering Problem}
\label{sec:statement:math}

Let $\mathcal{I} = \{0, \cdots, n-1 \}$ be an ordered set of $n$ target indices; the zero-based indexing matches the numerical implementation in~\Cref{sec:se:nlp}.
Let $\mathcal{Q}$ be a list of target positions $\mathbf{q}_i\in \mathbb{R}^3$ and $\mathcal{R}$ a list of rewards $r_i \in \mathbb{R}_{>0}$ for each $i \in \mathcal{I}$.
Let $\mathbf{p}_{\text{s}}, \mathbf{p}_{\text{e}}\in \mathbb{R}^3$ be the trajectory start and end positions, \Vmax~and \Amax~the maximum velocity and acceleration magnitudes, and $\Cmax\in \mathbb{R}_{>0}$ the maximum travel time.
The vehicle velocity is zero at $\mathbf{p}_{\text{s}}$ and $\mathbf{p}_{\text{e}}$.
The DVOP seeks a sequence $S^*$, a permutation of some subset of $\mathcal{I}$, that maximizes the collected reward $R^* = R(S^*) = \sum_{i \in S^*}r_i$, subject to the time-optimal trajectory through $S^*$ (from $\mathbf{p}_{\text{s}}$ to $\mathbf{p}_{\text{e}}$) not exceeding the travel budget~\Cmax.

Given a sequence $S$ with elements $s_i \in S$, the time-optimal travel cost $C(S)$ is formulated in~\cref{eq:statement:cost_function} based on the continuous dynamics~\eqref{eq:model:ode}.
The time variables $t_1 \tdots t_{\card{S}}$ are the visit times of targets $s_0 \tdots s_{\card{S}-1}$, $t_0$ the visit of~$\mathbf{p}_{\text{s}}$, and $t_{\card{S}+1}$ the visit of~$\mathbf{p}_{\text{e}}$.
\begin{equation}\label{eq:statement:cost_function}
\centering
   \begin{array}{lll}
      C(S)  =& \multicolumn{2}{l}{\displaystyle \min_{\textbf{a}, t} (t_{\card{S}+1} - t_0)} \\
      &\text{ s.t.} &\mathbf{p}(t_0) = \mathbf{p}_{\text{s}},~\mathbf{p}(t_{\card{S}+1}) = \mathbf{p}_{\text{e}},\\
      && \mathbf{v}(t_0) = \mathbf{0},~\mathbf{v}(t_{\card{S}+1}) = \mathbf{0},\\
      &&\mathbf{p}(t_i) = \mathbf{q}_{s_{i-1}}~\forall i = 1 \tdots \card{S}, \\
      &&t_{i-1} \leq t_i~\forall i = 1 \tdots \card{S}+1,\\
      &&\text{and~\cref{eq:model:ode,eq:constraints}}.
   \end{array}
\end{equation}
The DVOP is formulated in~\cref{eq:statement:formulation} as maximizing the reward collected by $S$ subject to the travel budget~\Cmax, where $\mathcal{S}$ is the set of all permutations of all subsets of $\mathcal{I}$.
\begin{equation}\label{eq:statement:formulation}
\centering
   \begin{array}{lll}
      R^*=&\multicolumn{2}{l}{\displaystyle  \max_{S\in\mathcal{S}} \sum_{i \in S} r_i,} \\
      &\textrm{ s.t.} & \displaystyle C(S) \leq \Cmax.
   \end{array}
\end{equation}

\section{Large Neighborhood Search heuristic solver}\label{sec:vns}



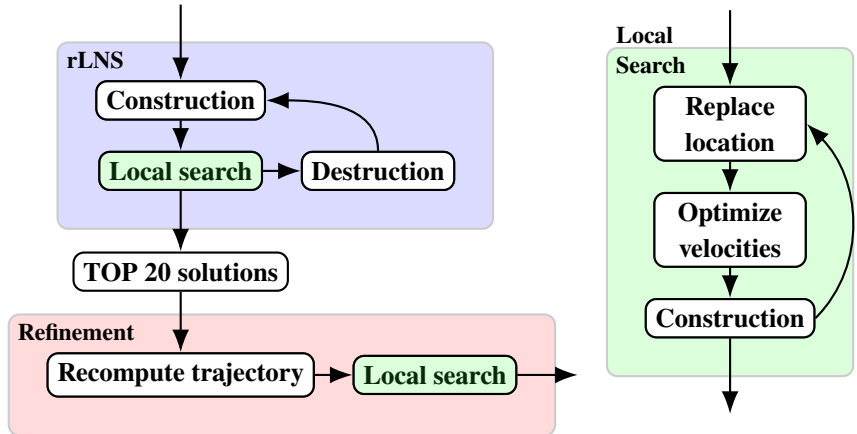
\begin{figure}[h]
    \centering
    \begin{tikzpicture}[node distance=0.4cm and 0.5cm]

        \node[block] (construction) {\textbf{Construction}};
        \node[block, below=of construction] (local) {\textbf{Local search}};
        \node[block, right=of local] (destruction) {\textbf{Destruction}};
        
        \node[container={rLNS}, fill=blue!70, opacity=0.2, fit=(construction)(local)(destruction)] (lnsbox) {};
        
        \node[block, fill=white!70] (construction) {\textbf{Construction}};
        \node[block, right=of local, fill=white!70] (destruction) {\textbf{Destruction}};
        \node[block, below=of construction, fill=white!70] (local) {\textbf{Local search}};
        \node[block, below=of construction, fill=green!70, opacity=0.2] (local) {\textbf{Local search}};
        
        \draw[arrow] (construction.south) -- (local.north);
        
        \draw[arrow] 
            (local.east) 
            to[out=0, in=180] (destruction.west);
        
        \draw[arrow]
            (destruction.north)
            to[out=90, in=0] (construction.east);
        
        \node[block, below=0.8cm of local] (top) {\textbf{TOP 20 solutions}};
        
        \node[block, below=0.8cm of top] (recompute) {\textbf{Recompute trajectory}};
        \node[block, right=of recompute] (ls2) {\textbf{Local search}};
        \node[container={Refinement}, fill=red!70, opacity=0.2, fit=(recompute)(ls2)] (lsbox) {};
        
        \node[block, below=0.8cm of top, fill=white!70] (recompute) {\textbf{Recompute trajectory}};
        \node[block, right=of recompute, fill=white!70] (ls2) {\textbf{Local search}};
        \node[block, right=of recompute, fill=green!70, opacity=0.2] (ls2) {\textbf{Local search}};
        
        \draw[arrow] (local.south) -- (top.north);
        \draw[arrow] (top.south) -- (recompute.north);
        \draw[arrow] (recompute.east) -- (ls2.west);
        \draw[arrow] ($(construction.north) + (0,1cm)$) -- (construction.north);
        \draw[arrow] (ls2.east) -- ($(ls2.east) + (0.8cm,0)$);

        \node[block, right=5cm of lnsbox.south] (velocity) {\textbf{Optimize}\\[0.5mm]\textbf{velocities}};
        \node[block, above=of velocity, fill=white!70] (replace) {\textbf{Replace}\\[0.5mm]\textbf{location}};
        \node[block, below=of velocity] (construction) {\textbf{Construction}};
        
        \node[container={}, fill=green!70, opacity=0.2, fit=(replace)(velocity)(construction),  label={[anchor=north west, yshift=4mm, font=\bfseries\footnotesize]
           north west:\parbox{1.6cm}{Local\\Search}}] (lsbox) {};
                        
        \node[block, above=of velocity, fill=white!70] (replace) {\textbf{Replace}\\[0.5mm]\textbf{location}};
        \node[block, right=5cm of lnsbox.south] (velocity) {\textbf{Optimize}\\[0.5mm]\textbf{velocities}};
        \node[block, below=of velocity, fill=white!70] (construction) {\textbf{Construction}};
        
        \draw[arrow] (replace.south) -- (velocity.north);
        \draw[arrow] (velocity.south) -- (construction.north);
        \draw[arrow] (construction.south) -- ($(construction.south) - (0,1cm)$);
        \draw[arrow] (construction.east) to[out=45, in=-45] (replace.east);
        \draw[arrow] ($(replace.north) + (0,1cm)$) -- (replace.north);
        
    \end{tikzpicture}
    \caption{Overview of the proposed heuristic: a two-stage process where random Large Neighborhood Search (rLNS) generates diverse candidate solutions, followed by refinement of the best found solutions.}
    \label{fig:heuristic_vis}
    \vspace{-1em}
\end{figure}

The proposed heuristic is intentionally lightweight to yield effective solutions (see~\cref{tbl:lns_bnb_comparison}) in real time.
It serves both as a fast standalone planner and as a means of accelerating the BnB solver by providing a tight upper bound on the optimal solution.
The method has two stages: a random Large Neighborhood Search (rLNS) metaheuristic~\citep{pisinger2019lns} for rapid generation of diverse candidate solutions, followed by a Local Search refinement of the best solutions, as visualized in~\cref{fig:heuristic_vis}.

To overcome the limitations of prior heuristic solvers~\citep{meyer2022kinematic}, which uniformly sample entry velocities at the targets, we introduce a greedy gradient-based optimization of entry velocities at each insertion.
Sparse sampling risks missing the time-optimal solution, while dense sampling enlarges the search space and increases the computational cost.
Our method removes the need for velocity sampling entirely by directly optimizing the affected trajectory segment with the point-mass planner for multi-waypoint trajectories~\citep{teissing2024real}.
Consequently, the solver performs the combinatorial optimization of target selection and ordering and the continuous optimization of entry velocities simultaneously.

\subsection{Trajectory representation}\label{sec:heuristic_trajectory}

As shown by~\cite{teissing2024real}, a time-optimal one-dimensional point-mass trajectory subject to~\Vmax~and~\Amax~comprises three segments: acceleration, cruise at constant velocity, and deceleration.
This \emph{bang-zero-bang} trajectory is represented by the following analytically solvable equations:
\begin{equation}\label{eq:pmm_1seg}
	\begin{aligned}
		p_1 &= p_s + v_s \Delta t_1 + \dfrac{1}{2} a_1 \Delta t_1^2,
		&v_1 &= v_s + a_1 \Delta t_1,\\
        p_2 &= p_1 + v_1 \Delta t_2,
        &v_2 &= v_1 = \Vmax,\\
		p_e &= p_2 + v_2 \Delta t_3 + \dfrac{1}{2} a_2 \Delta t_3^2,
		&v_e &= v_2 + a_2 \Delta t_3,
	\end{aligned}
\end{equation}
where \(p_s, v_s\) and \(p_e, v_e\) are the initial and terminal states of a single axis, and \(\Delta t_1, \Delta t_2, \Delta t_3 \in \mathbb{R}_{\geq 0}\) the segment durations.
The optimal control inputs $a_1, a_2$ satisfy
\begin{equation}\label{eq:pmm_acc_limits}
	a_i \in \{+\Amax^{\text{axis}}, -\Amax^{\text{axis}}\}, \quad i \in \{1,2\},
\end{equation}
where $\Amax^{\text{axis}}$ is the per-axis acceleration limit, equal to $\Amax \div \sqrt{d}$ during rLNS (\cref{sec:lns}) and axis-specific during refinement (\cref{sec:refinement}); it is the per-axis analogue of~\eqref{eq:accmag}.
If~\Vmax~is not reached, the cruise segment is omitted ($\Delta t_2=0$), leaving a \emph{bang-bang} trajectory.
Parameter $d$ represents spatial dimensionality.

\subsubsection{Limited thrust decomposition}\label{sec:ltd}

Extension to 3D requires synchronizing the per-axis durations.
As in~\cite{Foehn2022AlphaPilot, Romero2022PMMReplanningMPCC, teissing2024real}, we set the 3D duration \(T_{\mathrm{s}}\) to that of the slowest per-axis trajectory.
A per-axis scaling factor \mbox{$\gamma\in(0,1]$}, distinct for each axis, reduces the accelerations of faster axes to prolong their duration.
Substituting \(\gamma_i a_i\) for \(a_1\) and \(a_2\) in~\cref{eq:pmm_1seg} and enforcing \(T_{\mathrm{s}} = \Delta t_1 + \Delta t_2 + \Delta t_3\) adds one variable and one equation, so the problem remains analytically solvable~\citep{teissing2024real}.
The available acceleration~\Amax~is then only partially utilized:
\begin{equation}
    \lVert \mathbf{a}_{\gamma} \rVert \leq \Amax, \quad
    \mathbf{a}_{\gamma} = \begin{bmatrix} \gamma_x a_x \\   \gamma_y a_y \\ \gamma_z a_z \\ \end{bmatrix}, \quad
    \gamma_x, \gamma_y, \gamma_z \in (0,1],
\end{equation}
where $\mathbf{a}_{\gamma}$ is the per-axis scaled acceleration.
To maximize acceleration utilization and thus minimize \(T_{\mathrm{s}}\), the per-axis limits must be distributed unevenly while satisfying~\cref{eq:accmag,eq:velmag}.
Rather than solving the non-linear constraint in~\cref{eq:accmag} directly, \cite{teissing2024real} proposed the \emph{Limited Thrust Decomposition} (LTD) algorithm, which iteratively approximates an optimal acceleration distribution to minimize \(T_{\mathrm{s}}\).

LTD also accommodates external forces such as gravity $\mathbf{g}=[0,0,-g]^{\top}$.
Using the convention of~\cref{sec:statement:model}, the thrust acceleration $\mathbf{a}_{\text{th}}=\mathbf{a}-\mathbf{g}$ is distributed across axes, so~\eqref{eq:accmag} reads $\lVert \mathbf{a}_{\text{th}} \rVert \leq \Amax$ with $\Amax = \Fmax \div m$.

Although the resulting trajectory is only near-optimal, low computation time is crucial for the heuristic, and the trajectory is deterministic given the start state \((\mathbf{p}_{\text{s}}, \mathbf{v}_{\text{s}})\) and end state \((\mathbf{p}_{\text{e}}, \mathbf{v}_{\text{e}})\).

In the remainder of this section, $S=\langle s_0,\ldots,s_{\card{S}-1}\rangle$ denotes an ordered solution sequence of target indices $s_i\in\mathcal{I}$, as in~\cref{sec:statement:math}; it excludes the start and end positions.
The LTD-planner travel cost from $\mathbf{p}_{\text{s}}$ through $S$ to $\mathbf{p}_{\text{e}}$ is denoted $\tilde{C}(S)$.
Since the LTD planner restricts the trajectory to a synchronized per-axis bang-zero-bang structure, its optimum upper-bounds the true minimum-time cost $C(S)$ of~\cref{eq:statement:cost_function}:
\begin{equation}\label{eq:lns:ctildefeasible}
    \tilde{C}(S) \geq C(S).
\end{equation}
Hence any solution with $\tilde{C}(S)\leq \Cmax$ is feasible for the DVOP, and $\tilde{C}$ is the cost function used throughout the heuristic.

Each target $s_i$ in $S$ is associated with an entry velocity vector $\mathbf{v}_i\in\mathbb{R}^3$, and the pair $\sigma_i=(s_i,\mathbf{v}_i)$ is referred to as a \emph{visit}.
The full ordered sequence of positions visited by the trajectory is
\begin{equation}\label{eq:lns:path}
    P(S)=\big\langle\mathbf{p}_{\text{s}},\mathbf{q}_{s_0},\ldots,\mathbf{q}_{s_{\card{S}-1}},\mathbf{p}_{\text{e}}\big\rangle,
\end{equation}
where $\mathbf{p}_{\text{s}}$ and $\mathbf{p}_{\text{e}}$ have zero velocity and are never modified.
The visits $\sigma_i$ together with $P(S)$ fully describe the trajectory.

\subsection{Insertion procedure}\label{sec:insertion}

This section describes a greedy procedure for inserting a new target into the current solution, used by the \textit{Construction}, \textit{Local search}, and \textit{Replace location} methods in~\cref{fig:heuristic_vis}.
Given a sequence $S$ and a candidate target $s_{\text{new}} \notin S$, the goal is to find the index $i\in\{0,\ldots,\card{S}\}$ minimizing the cost of
$S_{\text{new}} = \langle s_0, \ldots, s_{i-1}, s_{\text{new}}, s_i, \ldots, s_{\card{S}-1} \rangle$,
where $i=0$ prepends and $i=\card{S}$ appends $s_{\text{new}}$.
The cost $\tilde{C}(S_{\text{new}})$ is computed with the LTD planner~\citep{teissing2024real}.
The candidate target $s_{\text{new}}$ is chosen by the method requesting insertion (construction, \cref{sec:construction}); the procedure below only finds the best insertion index and adjusts the entry velocities in $S_{\text{new}}$.

\begin{figure}[t]
    \centering
    \includegraphics[width=\linewidth]{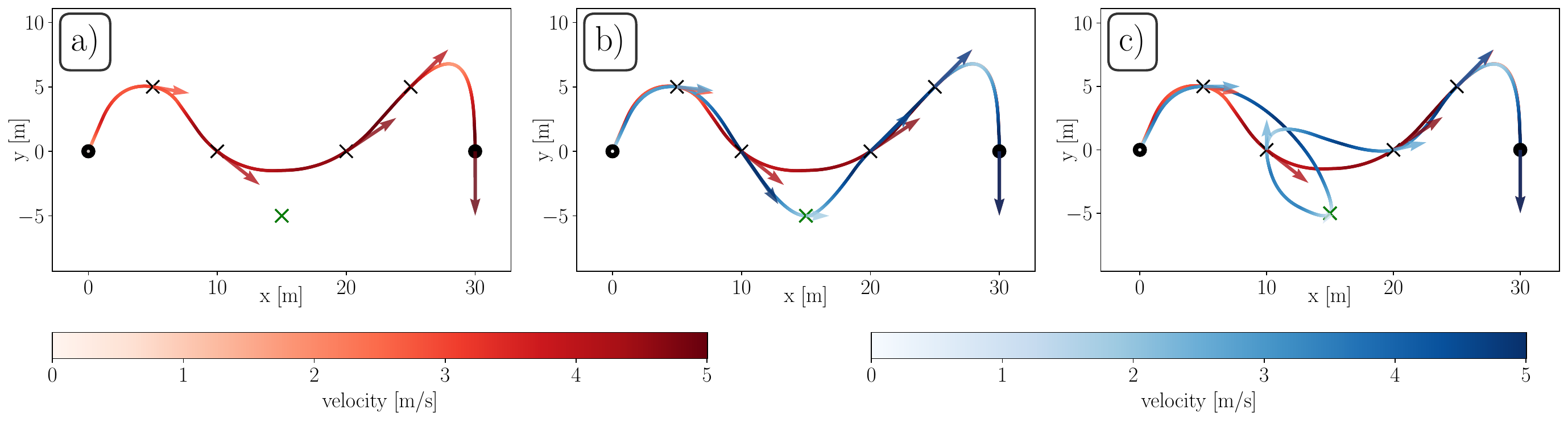}
    \vspace{-1.5em}
    \caption{Visualization of the insertion procedure. Black dots mark the fixed window boundaries $\pi_{\ell}$ and $\pi_{h}$, whose velocities are held constant (zero for the start $\mathbf{p}_{\text{s}}$ and end $\mathbf{p}_{\text{e}}$ positions). Velocities at positions marked with X are subject to optimization, together with the velocity at the newly inserted target $s_{\text{new}}$, marked with a green X. The local window before insertion, $W_{\text{old}}$ (a), is shown in red; the window after insertion, $W_{\text{new}}$, is colored blue, with arrows representing the entry velocities at each visit. In the presented case, insertion index (b) is selected over (c), as it yields a smaller cost increase.}
    \label{fig:insertion}
    \vspace{-1em}
\end{figure}

Inserting $s_{\text{new}}$ affects only a local window of $P(S)$.
Indexing $P(S)=\langle\pi_0,\ldots,\pi_{\card{S}+1}\rangle$ with $\pi_0=\mathbf{p}_{\text{s}}$, $\pi_j=\mathbf{q}_{s_{j-1}}$ for $1\leq j\leq\card{S}$, and $\pi_{\card{S}+1}=\mathbf{p}_{\text{e}}$, insertion at index $i$ places $\mathbf{q}_{s_{\text{new}}}$ between $\pi_i$ and $\pi_{i+1}$.
For each $i\in\{0,\ldots,\card{S}\}$, a window is extracted with bounds
\begin{equation}
    \ell = \max(0,\, i-\eta), \qquad h = \min(\card{S}+1,\, i+\eta),
\end{equation}
where $\eta$ is the neighborhood size that determines the local position window.
The old and new windows are the sub-sequences
\begin{equation}
    W_{\text{old}}(i) = \langle \pi_{\ell}, \ldots, \pi_{h} \rangle, \qquad
    W_{\text{new}}(i) = \langle \pi_{\ell}, \ldots, \pi_{i}, \mathbf{q}_{s_{\text{new}}}, \pi_{i+1}, \ldots, \pi_{h} \rangle.
\end{equation}
The boundary positions $\pi_{\ell}$ and $\pi_{h}$ are fixed, with velocities held at their current values (zero for $\mathbf{p}_{\text{s}}$ and $\mathbf{p}_{\text{e}}$), while the entry velocities of the interior visits of $W_{\text{new}}(i)$, including $\mathbf{q}_{s_{\text{new}}}$, are optimized by gradient descent.
Since segments outside the window depend only on the frozen boundary states, their costs are unaffected, and the insertion cost is fully captured by the window:
\begin{equation}
    \tilde{C}_{\text{ins}}(i) = \tilde{C}(W_{\text{new}}(i)) - \tilde{C}(W_{\text{old}}(i)).
\end{equation}
The optimal insertion index is
\begin{equation}
    i_{\text{best}} = \arg\min_i \tilde{C}_{\text{ins}}(i),
\end{equation}
subject to $\tilde{C}(S) + \tilde{C}_{\text{ins}}(i_{\text{best}}) \leq \Cmax$.
If no $i\in\{0,\ldots,\card{S}\}$ satisfies this, the insertion is unsuccessful.
As~\cref{fig:insertion} shows, entry velocities far from $s_{\text{new}}$ are only marginally affected, motivating optimization of the local window only.
We use a smaller neighborhood during rLNS ($\eta=2$) for speed and increase it to $\eta=3$ during refinement.

\subsection{Large neighborhood search}~\label{sec:lns}
The random Large Neighborhood Search (rLNS) component (blue block in~\cref{fig:heuristic_vis}) generates a diverse set of feasible solutions in a short time.
Our variant, summarized in~\cref{alg:lns}, comprises construction, local search, and destruction procedures (below).
A total of $N_{\text{iter}}=200$ iterations are run, retaining the best $N_{\text{keep}}=20$ unique solutions, where uniqueness is defined solely by the target sequence in $S$.
To increase diversity, a randomized greedy rule favors targets that have appeared less frequently, tracked by the counter $f$, incremented by one whenever a target enters a solution (line~\ref{line:counter}).
If no improvement occurs, the destruction ratio $\rho$ is increased (line~\ref{line:destruction}) and reset when a new best solution is found (line~\ref{line:reset}).
To reduce computational cost, fixed per-axis constraints are applied when generating point-mass trajectories, as in~\cite{meyer2022kinematic}.
For both 2D and 3D instances, the per-axis limits are $\frac{\Vmax}{\sqrt{d}}$ and $\frac{\Amax}{\sqrt{d}}$, with $d\in\{2,3\}$ indicating the problem's spatial dimension.

\begin{algorithm}[h]
    \caption{rLNS heuristic for DVOP}
    \begin{algorithmic}[1]
        \small
        \Statex \textbf{Input:} $\mathcal{I}$ - set of target indices, $N_{\text{keep}}$ - number of solutions to retain
        \Statex \textbf{Output:} $\mathcal{P}$ - sorted set of candidate solutions
        \algrule
        \State $\mathcal{P} \gets \emptyset$
        \State $f \gets \{\}$ \Comment{hash-table~\cref{eq:freq}}
        \For{each $i$ in $\mathcal{I}$}
            \State $f(i) \gets 1$
        \EndFor
        \State $\rho \gets 0.1$ \Comment{destruction ratio, see \cref{sec:destruction}}
        \State $S \gets \langle\,\rangle$ \Comment{empty sequence; trajectory is $\mathbf{p}_{\text{s}}\to\mathbf{p}_{\text{e}}$}
        \For{$k = 1, \ldots, N_{\text{iter}}$}
            \State $S \gets$ \textbf{construction}$(S, f) \qquad$ \Comment{\cref{alg:construction}} \label{line:construction}
            \State $S \gets$ \textbf{replaceLocation}$(S, f) \;$ \Comment{\cref{alg:replace_location}} \label{line:ls:replace}
            \State $S \gets$ \textbf{optimizeVelocities}$(S)$ \Comment{\cite{teissing2024real}} \label{line:ls:optimize}
            \State $S \gets$ \textbf{construction}$(S, f) \qquad$ \Comment{\cref{alg:construction}} \label{line:ls:construction}

            \If{$R(S) > \displaystyle \max_{S'\in\mathcal{P}}R(S')$}
                \State $\rho \gets 0.1$ \label{line:reset}
            \Else
                \State $\rho \gets \min(\rho + 0.1, 0.8)$ \label{line:destruction}
            \EndIf
            \For{each $s \in S$}
                \State $f(s) \gets f(s) + 1$ \label{line:counter}
            \EndFor
            \If{$S \notin \mathcal{P}$}
                \State $\mathcal{P} \gets \mathcal{P} \cup \{S\}$ \Comment{retain unique solution}
            \EndIf
            \State $S \gets$ \textbf{destruction}$(S, d) \qquad$ \Comment{\cref{sec:destruction}} \label{line:lns:destroy}
        \EndFor
        \State \textbf{return} $N_{\text{keep}}$ best solutions in $\mathcal{P}$
    \end{algorithmic}
    \label{alg:lns}
\end{algorithm}

\subsubsection{Construction}\label{sec:construction}

Each rLNS iteration begins with a construction phase in which targets are selected at random for insertion.
Each target $s_i \notin S$ is assigned a randomized weight
\begin{equation}\label{eq:freq}
    w(s_i) = \frac{\log u_i}{f(s_i)}, \quad u_i \sim \mathrm{Uniform}(0,1) \text{ i.i.d.,}
\end{equation}
and the target
\begin{equation}\label{eq:selection}
    s_{\text{add}} = \displaystyle \argmin_{s\in \mathcal{I}\setminus S} w(s)
\end{equation}
with the lowest weight is selected.
It is greedily inserted at the index and entry velocity that minimize the cost increase (\cref{sec:insertion}).
If insertion fails, the next-lowest-weight target is selected.
Construction ends when all targets are inserted or no remaining target fits the budget~\Cmax.
The procedure is summarized in~\cref{alg:construction}.
\begin{algorithm}[h]
    \caption{Construction procedure}
    \begin{algorithmic}[1]
        \small
        \Statex \textbf{Input:} $S$ - current solution, $\mathcal{I}$ - set of all targets, $f$ - frequency hash-table
        \Statex \textbf{Output:} $S$ - new solution
        \algrule
        \State inserted $\gets$ true
        \While{inserted}
            \State inserted $\gets$ false
            \State $U \gets \mathcal{I} \setminus S$
            \For{each $s \in U$}
                \State $w(s) \leftarrow \frac{\log u_s}{f(s)}$ with $u_s\sim\mathrm{Uniform}(0,1)$ \Comment{\cref{eq:freq}}
            \EndFor
            \While{$U \neq \emptyset$}
                \State $s_{\text{add}} \leftarrow \argmin_{s\in U} w(s)$ \Comment{\cref{eq:selection}}
                \State $U \leftarrow U \setminus \{s_{\text{add}}\}$
                \State $S_{\text{new}} \gets$ \textbf{insertion}$(S, s_{\text{add}})$ \Comment{\cref{sec:insertion}}
                \If{$\tilde{C}(S_{\text{new}}) \leq \Cmax$}
                    \State $S \gets S_{\text{new}}$
                    \State inserted $\gets$ true
                    \State \textbf{break}
                \EndIf
            \EndWhile
        \EndWhile
        \State \textbf{return} $S$
    \end{algorithmic}
    \label{alg:construction}
\end{algorithm}

\subsubsection{Local Search}~\label{sec:local_search}
During the local search phase, a neighborhood of the current solution $S$ is explored to increase reward or reduce cost via \textit{replace location} and \textit{optimize velocities}, followed by \textit{construction} to maximize the collected reward.

The \textit{replace location} procedure (line~\ref{line:ls:replace} in~\cref{alg:lns}) removes a target $s_i$, starting from $i=0$.
Since $\mathbf{p}_{\text{s}}$ and $\mathbf{p}_{\text{e}}$ are not in $S$, every target is removable.
Construction (\cref{sec:construction}) then refills the solution with $s_i$ blocked.
If the resulting $S_{\text{new}}$ is no better than $S$, the change is discarded and $s_{i+1}$ is removed next, continuing over all $i \in \{0,\ldots,\card{S}-1\}$.
Whenever $S_{\text{new}}$ has higher reward, or equal reward and lower cost, we set $S \leftarrow S_{\text{new}}$, reset $i=0$, and restart.
The pseudocode is in~\cref{alg:replace_location}.

The \textit{optimize velocities} procedure (line~\ref{line:ls:optimize} in~\cref{alg:lns}) applies the gradient-based method of~\cite{teissing2024real} to reduce the cost of $S$.
Although it cannot increase the reward, lower cost may enable further insertions, so it is followed by another \textit{construction} phase (line~\ref{line:ls:construction} in~\cref{alg:lns}).
\begin{algorithm}[h]
    \caption{Replace location procedure}
    \begin{algorithmic}[1]
        \small
        \Statex \textbf{Input:} $S$ - current solution, $f$ - frequency hash-table
        \Statex \textbf{Output:} $S$ - new solution
        \algrule
        \State improvement $\gets$ true
        \While{improvement}
            \State improvement $\gets$ false
            \For{each $s \in S$}
                \State $S_{\text{tmp}} \gets$ \textbf{remove}$(s)$ from $S$
                \State $S_{\text{new}} \gets$ \textbf{construction}$(S_{\text{tmp}}, f)$ with $s$ blocked \Comment{\cref{alg:construction}}
                \If{$R(S_{\text{new}}) > R(S)$ \textbf{or} ($R(S_{\text{new}}) = R(S)$ \textbf{and} $\tilde{C}(S_{\text{new}}) < \tilde{C}(S)$)}
                    \State $S \gets S_{\text{new}}$
                    \State improvement $\gets$ true
                    \State \textbf{break}
                \EndIf
            \EndFor
        \EndWhile
        \State \textbf{return} $S$
    \end{algorithmic}
    \label{alg:replace_location}
\end{algorithm}

\subsubsection{Destruction}\label{sec:destruction}

Two heuristics, which are selected at random on each iteration, partially destroy the current solution $S$ (line~\ref{line:lns:destroy} in~\cref{alg:lns}).
The first removes $D$ randomly selected individual targets independently; the second removes a random subsequence of $D$ consecutive targets.
Since $\mathbf{p}_{\text{s}}$ and $\mathbf{p}_{\text{e}}$ are not in $S$, they are never removed.
The number of targets removed is
\begin{equation}
    D = \left\lceil \rho \cdot \card{S} \right\rceil,
\end{equation}
where $\rho$ starts at $10\%$, increases by $10\%$ per iteration without improvement up to $80\%$, and resets to $10\%$ when a new best solution is found.

\subsection{Solution refinement with acceleration magnitude constraints}\label{sec:refinement}

The $N_{\text{keep}}=20$ best unique rLNS solutions are each further refined.
First, their trajectories are recomputed with the LTD approach (\cref{sec:ltd}) under the magnitude constraints~\Amax~and~\Vmax, instead of the conservative per-axis limits $\frac{\Vmax}{\sqrt{d}}$, $\frac{\Amax}{\sqrt{d}}$.
This yields an average cost reduction of 10--\SI{20}{\percent}, allowing additional insertions, and is followed by a local search akin to~\cref{sec:local_search}.
In \textit{construction}, rather than random selection, all insertions are evaluated, and the one with the highest reward-to-cost-increase ratio is chosen.
The process repeats for each selected solution until five consecutive iterations yield no improvement, with up to 30\% of the solution destroyed per iteration, down from 80\% in~\cref{sec:lns}.

\section{Branch and Bound solver}\label{sec:se}

We propose a BnB solver that obtains exact DVOP solutions by using an MILP relaxation of the vehicle dynamics, based on trajectory primitives through target triplets, to compute reward upper bounds during the search.
Because of the vehicle dynamics, reward collection is intertwined with time-optimal trajectory planning (\cref{eq:statement:formulation,eq:statement:cost_function}); unlike the OP, such a problem cannot be formulated as a MILP and solved to optimality by an off-the-shelf solver.
Moreover, no exact time-optimal point-mass planner exists for the non-convex cost $C(S)$ in~\cref{eq:statement:cost_function}.
We therefore propose a hybrid procedure combining the NLP formulation of~\cref{eq:statement:cost_function}, denoted $C^{\text{NLP}}(S)$ (\cref{eq:nlp:c_full}), to obtain locally optimal $C(S)$, and an MILP-based relaxation of $C(S)$ to obtain the reward upper bound for each sequence $S$.
The exactness of the BnB solution relies directly on the exactness of the solutions to~\cref{eq:statement:cost_function} and its relaxations~\cref{eq:se:crel,eq:se:crelhat} (\cref{sec:se:bounds}).
The procedure is described in~\cref{sec:se:proc}, the NLP cost functions in~\cref{sec:se:nlp}, and the MILP relaxation in~\cref{sec:se:mip}.

\subsection{Bounds and relaxations}\label{sec:se:bounds}
During solution-tree exploration, each node is uniquely defined by its sequence $S\in \mathcal{S}$, the set of all sequences optimized upon (\cref{sec:statement}).
A node is thus referred to simply as a sequence, although in implementation it also stores its reward upper bound $R_{\text{UB}}$ and priority $p_{\text{o}}$.
A node $S$ is branched by appending each unused target from $\mathcal{U}(S) = \{ i \in \mathcal{I} : i \notin S\}$ to the end of $S$, producing $\card{\mathcal{U}}$ leaves of length $\card{S}+1$.
A sequence $S$ is valid if its cost $C(S)$ satisfies the travel budget; any valid sequence may be the optimum $S^*$.
The reward collected by $S$ is
\begin{equation}
        R(S) = \textstyle \sum_{i = 0}^{\card{S}-1}r_{s_i}.\label{eq:se:r}
\end{equation}
To search efficiently, we seek the reward upper bound $R_{\text{UB}}(S)$, an estimate of the reward obtainable by traveling from $\mathbf{p}_{\text{s}}$ (zero initial velocity) through $S$ in order, then through any permutation of any subset of $\mathcal{U}(S)$, ending at $\mathbf{p}_{\text{e}}$ (zero velocity).
Define the bounding sequence $\hat{S}(S)$, which begins with the last element of $S$ and continues with a permutation of a subset of targets not in $S$ that maximizes $R_{\text{UB}}(S)$.
Its reward is
\begin{equation}
    \hat{R}(\hat{S}) = \textstyle \sum_{i = 1}^{\card{\hat{S}}-1}r_{\hat{s_i}},\label{eq:se:rhat}
\end{equation}
where the sum starts at $i=1$ because $\hat{s}_0$ is the last element of $S$, already counted in $R(S)$.
The reward upper bound is
\begin{equation}
        R_{\text{UB}}(S) = R(S) + \hat{R}(\hat{S}).\label{eq:se:rub}
\end{equation}
Naively, $\hat{S}(S)$ would be the last element of $S$ followed by any permutation of targets not in $S$; we seek a mapping $S \rightarrow \hat{S}(S)$ that also respects the travel budget and yields a tighter bound.
Let $\hat{T}_{\max}(S)$ be the remaining budget after traveling from $\mathbf{p}_{\text{s}}$ (zero velocity) through $S$:
\begin{equation}
\hat{T}_{\max}(S) = \Cmax - C_{\text{rel}}(S),\label{eq:se:that}
\end{equation}
where $C_{\text{rel}}(S)$ relaxes $C(S)$ by removing the constraints to visit $\mathbf{p}_{\text{e}}$ and end with zero velocity:
\begin{equation}\label{eq:se:crel}
\centering
   \begin{array}{lll}
      C_{\text{rel}}(S) =  & \multicolumn{2}{l}{\displaystyle \min_{\mathbf{a}, t} (t_{\card{S}} - t_0),} \\
      &\text{ s.t.} &\mathbf{p}(t_0) = \mathbf{p}_{\text{s}},~\mathbf{v}(t_0) = 0,\\
      &&\mathbf{p}(t_i) = \mathbf{q}_{s_{i-1}}~\forall i = 1 \ldots \card{S},\\
      &&t_{i-1} \leq t_i~\forall i = 1 \ldots \card{S},\\
      &&\text{and~\cref{eq:model:ode,eq:constraints}}.
   \end{array}
\end{equation}
The variables $t_0 \ldots t_{\card{S}}$ are the visit times of $\mathbf{p}_{\text{s}}$ and the targets $s_i\in S$ in order.
For any $S$,~\cref{eq:se:crelneq} holds, so $C_{\text{rel}}(S)$ is a lower bound on $C(S)$ and $\hat{T}_{\max}(S)$ an upper bound on the remaining budget:
\begin{equation}
    C_{\text{rel}}(S) \leq C(S).\label{eq:se:crelneq}
\end{equation}
\noindent
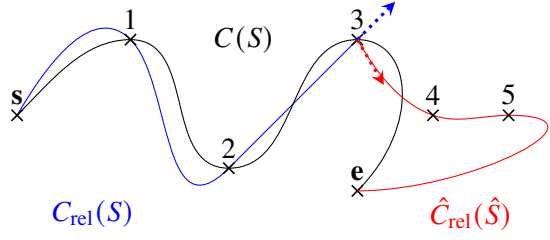
\begin{figure}
    \centering
\begin{tikzpicture}[tangent/.style={%
    in angle={(180+#1)},Hobby finish,
    designated Hobby path=next,out angle=#1},
    pics/vert/.style={code={
    \draw[solid,semithick,black] (0,0) pic{cross=3pt} node[above]{#1} ;}}]
\draw [use Hobby shortcut, blue] 
   ([out angle=60]0,0) pic{vert=$\mathbf{s}$}  .. ([tangent=-45]1.5,1) pic{vert=1}
    ..  ([tangent=45]2.8,-0.7) pic{vert=2} .. ([in angle=225]4.5,1) pic{vert=3};
\draw [use Hobby shortcut] 
   ([out angle=45]0,0) pic{}  .. ([tangent=0]1.5,1) pic{}
    ..  ([tangent=0]2.8,-0.7) pic{} .. ([tangent=0]4.5,1) pic{} .. ([in angle=45]4.5,-1) pic{vert=$\mathbf{e}$};
\draw [use Hobby shortcut,red] 
   ([out angle=-60] 4.5,1) pic{} .. ([tangent=-20] 5.5,0) pic{vert=4}  .. ([tangent=0]6.5,0) pic{vert=5}
    ..  ([tangent=180]4.5,-1) pic{};
\draw [very thick, dotted, blue, ->] (4.5,1) -- (5,1.5);
\draw [very thick, dotted, red, ->] (4.5,1) -- (4.85,0.4);
\node[blue] at (1,-1.3) {$C_{\text{rel}}(S)$};
\node[red] at (6,-1.3) {$\hat{C}_{\text{rel}}(\hat{S})$};
\node[] at (3,1) {${C}(S)$};
\end{tikzpicture}
    \caption{Illustration of time-optimal trajectories for cost functions $C(S)$, $C_{\text{rel}}(S)$ and $\hat{C}_{\text{rel}}(\hat{S})$ with given example sequence $S = \langle 1,2,3 \rangle$ and its bounding sequence $\hat{S} = \langle 3,4,5 \rangle$.}
    \label{fig:example_cost_paths}
    \vspace{-1em}
\end{figure}
Define $\hat{C}_{\text{rel}}(\hat{S})$ as a relaxation of $C(\hat{S})$ that removes the constraints to visit $\mathbf{p}_{\text{s}}$ and start with zero velocity:
\begin{equation}\label{eq:se:crelhat}
\centering
   \begin{array}{lll}
      \hat{C}_{\text{rel}}(\hat{S}) =  &\multicolumn{2}{l}{\displaystyle \min_{\mathbf{a}, t} (t_{\card{\hat{S}}} - t_{0}),} \\
      &\text{ s.t.} &\mathbf{p}(t_{\card{\hat{S}}}) = \mathbf{p}_{\text{e}},~\mathbf{v}(t_{\card{\hat{S}}}) = \mathbf{0},\\
      &&\mathbf{p}(t_i) = \mathbf{q}_{\hat{s}_{i}}~\forall i = 0 \tdots \card{\hat{S}}-1, \\
      &&t_{i-1} \leq t_i~\forall i = 1 \tdots \card{\hat{S}},\\
      &&\text{and~\cref{eq:model:ode,eq:constraints}}.
   \end{array}
\end{equation}
The variables $t_0 \ldots t_{\card{\hat{S}}-1}$ are the visit times of targets in $\hat{S}$, and $t_{\card{\hat{S}}}$ the visit of $\mathbf{p}_{\text{e}}$ (with zero velocity).
Like $C_{\text{rel}}(S)$, $\hat{C}_{\text{rel}}(\hat{S})$ is a lower bound on $C(\hat{S})$, so~\cref{eq:se:chatrelneqc} holds for any $\hat{S}$:
\begin{equation}
        \hat{C}_{\text{rel}}(\hat{S}) \leq C(\hat{S}).\label{eq:se:chatrelneqc}
\end{equation}
For any $S$ and bounding sequence $\hat{S}$ of members of $\mathcal{U}(S)$, with $S\hat{S}$ their concatenation,~\cref{eq:se:costsplusneq} also holds:
\begin{equation}
    C_{\text{rel}}(S) + \hat{C}_{\text{rel}}(\hat{S}) \leq C(S\hat{S}).\label{eq:se:costsplusneq}
\end{equation}
\Cref{fig:example_cost_paths} illustrates the relationship of the cost functions.
The sum $C_{\text{rel}}(S) + \hat{C}_{\text{rel}}(\hat{S})$ is a relaxation of $C(S\hat{S})$ in which the first-order continuity constraint on the vehicle velocity at the last target of $S$ is removed.
Let $\hat{\mathcal{S}}(S)$ be the set of all sequences beginning with the last element of $S$ and continuing with all permutations of all subsets of $\mathcal{U}(S)$, i.e., all bounding sequences of $S$.
The function $R^*_{\hat{C}_{\text{rel}}}(S)$ defines the maximal reward obtainable by any $\bar{S}\in\hat{\mathcal{S}}(S)$ subject to $\hat{C}_{\text{rel}}(\bar{S}) \leq \hat{T}_{\max}(S)$.
\begin{equation}\label{eq:se:rhatcrelopt}
\centering
   \begin{array}{ll}
    \displaystyle \hat{R}^*_{\hat{C}_{\text{rel}}}(S) = &\displaystyle \max_{\bar{S} \in \hat{\mathcal{S}}(S)} \hat{R}(\bar{S})\\
    &\text{ s.t. } \hat{C}_{\text{rel}}(\bar{S})\leq \hat{T}_{\max}(S)
   \end{array}
\end{equation}
Solving~\cref{eq:se:rhatcrelopt} yields the tightest upper bound $R_{\text{UB}}(S)$ and the mapping $S\to\hat{S}(S)$ is then defined by~\cref{eq:se:inalub} as:
\begin{equation}
    R_{\text{UB}}(S) = R(S) + \hat{R}(\hat{S}),\text{ where }\hat{R}^*_{\hat{C}_{\text{rel}}}(S)=\hat{R}(\hat{S}).\label{eq:se:inalub}
\end{equation}
The problem in~\cref{eq:se:rhatcrelopt} is comparable in complexity to the DVOP~\eqref{eq:statement:formulation}, as both optimize over a subset of $\mathcal{S}$, and $\hat{C}_{\text{rel}}(S)$ is a strong relaxation of $C(S)$ that removes the start-position and initial-velocity constraints but retains the non-convexity.
However, any $\hat{C}(S)$ relaxing $\hat{C}_{\text{rel}}(S)$ provides a valid lower bound, and the corresponding upper bound $\hat{R}^*_{\hat{C}}$ in~\cref{eq:se:rhatcopt} is valid if~\cref{eq:se:inversecr} holds:
\begin{equation}\label{eq:se:rhatcopt}
\centering
   \begin{array}{ll}
    \displaystyle \hat{R}^*_{\hat{C}}(S) = &\displaystyle \max_{\bar{S}\in \mathcal{S}} \hat{R}(\bar{S})\\
    &\text{ s.t. } \hat{C}(\bar{S})\leq \hat{T}_{\max}(S),
   \end{array}
\end{equation}
\begin{equation}\label{eq:se:inversecr}
   \hat{R}^*_{\hat{C}}(S) \geq  \hat{R}^*_{\hat{C}_{\text{rel}}}(S)\text{ if } \hat{C}(\bar{S}) \leq \hat{C}_{\text{rel}}(\bar{S})~\forall~\bar{S}\in\mathcal{S}.
\end{equation}
We therefore propose a novel MILP relaxation satisfying~\cref{eq:se:inversecr} in~\cref{sec:se:mip}, with a cost function $\hat{C}(S)$.

\subsection{Solution procedure}\label{sec:se:proc}
\begin{algorithm}[h]
\caption{BnB procedure for the DVOP}
\begin{algorithmic}[1]
\small
\Statex \textbf{Input:} $\mathcal{I}$, $\mathcal{Q}$, $\mathcal{R}$, $\mathbf{p}_{\text{s}}$, $\mathbf{p}_{\text{e}}$, $\Cmax$, $\Vmax$, $\Amax$
\Statex \textbf{Output:} $R^*, S^*$
\algrule
\State $O \gets \langle \langle S_0 = \langle\,\rangle,\, R_{\text{UB}}=0,\, p_{\text{o}}=0\rangle\rangle$
\State $R_{\text{LB}} \gets -1$
\State $S^* \gets \langle\,\rangle$
\While{$O$ is not empty}
    \State $\langle S_{\text{c}}, R_{\text{UB}},\,\cdot\,\rangle \gets \text{pop}(O)$
    \If{$R_{\text{UB}} > R_{\text{LB}}$}
    \State $C \gets C^{\text{NLP}}(S_{\text{c}})$ \Comment{\cref{eq:statement:cost_function,eq:nlp:c_full}}
    \If{$C \leq \Cmax$}
    \State $R \gets R(S_{\text{c}})$
    \Comment{\cref{eq:se:r}}
    \If{$R > R_{\text{LB}}$}
        \State $R_{\text{LB}} \gets R$
        \State $S^* \gets S_{\text{c}}$
    \EndIf
    \For{each target $i \in \mathcal{I} $ where $i \notin S_{\text{c}}$}
    \State $S \gets S_{\text{c}} \cup \{i\}$
    \State $C_{\text{rel}} \gets C_{\text{rel}}^{\text{NLP}}(S)$\Comment{\cref{eq:se:crel,eq:nlp:c_rel}}
    \State $\hat{S}, \hat{C} \gets \text{MILP}(S,C_{\text{rel}})$ \Comment{\Cref{sec:se:mip}}
    \State $R_{\text{UB}} \gets R_{\text{UB}}(S,\hat{S})$
    \Comment{\cref{eq:se:rub}}
    \If{$R_{\text{UB}} > R_{\text{LB}}$}
    \State $\hat{C}_{\text{rel}} \gets \hat{C}_{\text{rel}}^{\text{NLP}}(\hat{S})$
    \Comment{\cref{eq:se:crelhat,eq:nlp:c_hat_rel}}
    \State $p_{\text{o}} \gets p_{\text{o}}(S)$
    \Comment{\cref{eq:se:p}}
    \State $O \gets O \cup \langle S,R_{\text{UB}},p_{\text{o}}\rangle$
    \EndIf
    \EndFor
    \EndIf
    \EndIf
\EndWhile
\State $R^* \gets R_{\text{LB}}$
\end{algorithmic}
\label{alg:exact}
\end{algorithm}

The BnB procedure is summarized in~\cref{alg:exact}.
We initialize an ordered queue $O$ with the root node, the empty sequence $S_0 = \langle\,\rangle$, and the reward lower bound $R_{\text{LB}} = -1$.
While $O$ is non-empty, candidate sequences $S_{\text{c}}$ are popped by priority $p_{\text{o}}$.
If $S_{\text{c}}$ is a valid sequence whose upper bound $R_{\text{UB}}(S_{\text{c}})$ can still improve $R_{\text{LB}}$, we branch it into leaves $S = S_{\text{c}} \cup \{i\}$ for each $i\in \mathcal{I} : i \notin S_{\text{c}}$.
If $R(S_{\text{c}}) > R_{\text{LB}}$, the lower bound is updated.
$R_{\text{UB}}$ is given by~\cref{eq:se:rub}.
To bound the remaining budget after $S$, with possible visits to targets not in $S$, we use the end-point relaxed cost $C_{\text{rel}}$ of~\cref{eq:se:crel}.
If $R_{\text{UB}}(S) > R_{\text{LB}}$, $S$ is inserted into $O$ with its bound and priority $p_{\text{o}}$ (\cref{eq:se:p}), where $\hat{S}$ and $\hat{C}$ are the bounding sequence and MILP-relaxation cost from solving~\cref{eq:se:rhatcopt}, and $\hat{C}_{\text{rel}}(\hat{S})$ is the minimum-time cost from~\cref{eq:nlp:c_hat_rel}:
\begin{equation}
    p_{\text{o}}(S) = R(S) + \hat{R}(\hat{S}) \frac{\hat{C}(\hat{S})}{\hat{C}_{\text{rel}}(\hat{S})}.\label{eq:se:p}
\end{equation}
Since we maximize reward, $R_{\text{UB}}$ would conventionally order the queue.
Instead, we order by the heuristic $p_{\text{o}}$, exploiting that the budget used by a sequence is proportional to its collected reward.
We estimate the tightness of the MILP upper bound as the ratio of its cost lower bound $\hat{C}(\hat{S})$ to the minimum-time cost $\hat{C}_{\text{rel}}(\hat{S})$ to improve convergence.

\subsection{Mixed integer programming formulation of reward upper bound}\label{sec:se:mip}
To determine the node upper bounds, we propose a novel MILP relaxing the problem in~\cref{eq:se:rhatcrelopt}.
Standard VRP MILPs are graph-based, with edge costs determined by a property of target pairs (e.g., Euclidean distance, flight time, or energy), and decision variables selecting edges.
Since the DVOP budget is the maximal travel time, a second-order vehicle with limited acceleration would have to stop at each target to maintain velocity continuity, under-actuating the dynamics.
But because we seek a reward upper bound (a cost lower bound, \cref{eq:se:crelhat}), we may neglect the zero-velocity constraints between targets and assume maximal velocity~\Vmax.
We instead propose a tighter lower-bound cost where decision variables select time-optimal trajectory primitives through target triplets, since three targets is the minimum that may require a velocity change and hence acceleration, reflecting~\cref{eq:accmag}.

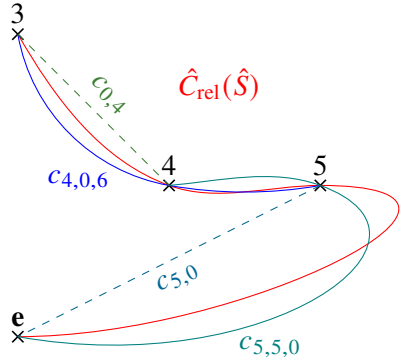
\begin{figure}[h]
    \centering
\begin{tikzpicture}[
    scale=0.6667,
    tangent/.style={%
        in angle={(180+#1)},Hobby finish,
        designated Hobby path=next,out angle=#1
    },
    pics/vert/.style={code={
        \draw[solid,semithick,black] (0,0) pic{cross=3pt}
        node[above]{#1} ;
    }}
]
    
\draw [use Hobby shortcut,red] 
   ([out angle=-60] 0,3) pic{vert=3} .. ([tangent=-20] 3,0) pic{vert=4}  .. ([tangent=-0]6,0) pic{vert=5} ..  ([tangent=180]0,-3) pic{vert=$\mathbf{e}$};
    
\draw [use Hobby shortcut,blue] 
   ([out angle=-80] 0,3) pic{} .. ([tangent=-10] 3,0) pic{}  .. ([in angle=-170]6,0) pic{};

\draw [use Hobby shortcut,teal] 
   ([out angle=5] 3,0) pic{}  .. ([tangent=-20]6,0) pic{} ..  ([tangent=170]0,-3) pic{};
   
\draw [dashed, OliveGreen] (0,3) -- (3,0) node [midway,sloped,above](c04) {$c_{0,4}$};
\draw [dashed, MidnightBlue] (6,0) -- (0,-3) node [midway,sloped,below](c5e) {$c_{5,0}$};
   
\node[red] at (4,2) {$\hat{C}_{\text{rel}}(\hat{S})$};
\node[blue] at (1.2,0.1) {$c_{4,0,6}$};
\node[teal] at (5,-3.2) {$c_{5,5,0}$};

\end{tikzpicture}
    \caption{Illustration of paths taken by time-optimal trajectory primitives used for MIP cost $\hat{C}$ and the trajectory of start-relaxed cost function $\hat{C}_{\text{rel}}$ through bounding sequence $\hat{S} = \langle 3, 4, 5 \rangle$.}
    \label{fig:example_lower_bound}
    \vspace{-1em}
\end{figure}
For a sequence $S$ seeking the upper bound, let $u_i \in \mathcal{U}(S)$ be the targets unvisited by $S$, $n^{\text{u}}=\card{\mathcal{U}}$ their count, $\mathcal{R}^{\text{u}} = \{ r_i \in \mathcal{R} : i \in \mathcal{U}\}$ the uncollected rewards $r^{\text{u}}_i$, and $\mathcal{Q}^{\text{u}} = \{ \mathbf{q}_i \in \mathcal{Q} : i \in \mathcal{U}\}$ the unvisited positions $\mathbf{q}^{\text{u}}_i$.
The decision matrix $\mathbf{X} \in \mathbb{Z}_{0,1}^{n^{\text{u}}\times (n^{\text{u}}+1) \times (n^{\text{u}}+1)}$ holds binary variables $x_{i,j,k}$, where $x_{i,j,k}=1$ indicates visiting $u_i$ via a primitive from $u_{j-1}$ to $u_{k-1}$; index $0$ for $j$ or $k$ denotes a primitive starting at the last target of $S$ ($\mathbf{q}_{s_{\card{S}-1}}$) or ending at $\mathbf{p}_{\text{e}}$ (with zero velocity).
The cost matrix $\mathbf{C} \in \mathbb{R}_{\geq 0}^{n^{\text{u}}\times (n^{\text{u}}+1) \times (n^{\text{u}}+1)}$ holds elements $c_{i,j,k}$, which are the time-optimal costs of the primitives through the positions indexed by $j$, $i$, $k$, defined in~\cref{eq:se:cprimitive} and implemented as an NLP in~\cref{eq:nlp:c_primitive}.
The variables $t_j$, $t_i$, $t_k$ are the respective visit times.
The index notation is illustrated in~\Cref{fig:example_lower_bound}.
\begin{equation}\label{eq:se:cprimitive}
\centering
   \begin{array}{lll}
      c_{i,j,k} = & \multicolumn{2}{l}{\displaystyle \min_{\mathbf{a}, t} (t_{k} - t_{j})} \\
      & \text{s.t.} &\mathbf{p}(t_j) =
      \begin{cases}
        \mathbf{q}_{s_{\card{S}-1}} & \text{if } j=0\\
        \mathbf{q}^{\text{u}}_{j-1} &\text{otherwise}
      \end{cases}\\[10pt]
      &&\mathbf{p}(t_{i}) = \mathbf{q}^{\text{u}}_{i},\\[2pt]
      &&\mathbf{p}(t_{k}) =
      \begin{cases}
      \mathbf{p}_{\text{e}} & \text{if }k=0\\
      \mathbf{q}^{\text{u}}_{k-1}&\text{otherwise}
      \end{cases}\\[10pt]
      && \mathbf{v}(t_k) = 0 \text{ if } k=0\\
      && t_j \leq t_i \leq t_k \text{ and \cref{eq:model:ode,eq:constraints}}
   \end{array}
\end{equation}
As the primitives on the edges of the bounding sequence do not overlap, additional costs $c_{0,i}$ (\cref{eq:c0i}) and $c_{i,0}$ (\cref{eq:ci0}) are incurred.
The cost $c_{0,i}$ is a lower bound on the travel time from the last target of $S$ to $u_i$, and $c_{i,0}$ a lower bound from $u_i$ to $\mathbf{p}_{\text{e}}$, where the vehicle stops.
\begin{equation}
    c_{0,i} = \frac{\norm{\mathbf{q}_{s_{\card{S}-1}}-\mathbf{q}^{\text{u}}_{i}}}{\Vmax}\label{eq:c0i}
\end{equation}
\begin{equation}\label{eq:ci0}
\displaystyle c_{i,0} =
    \begin{cases}
    \displaystyle\sqrt{\frac{2\norm{\mathbf{q}_i^{\text{u}}-\mathbf{p}_{\text{e}}}}{\Amax}} \text{ if } \displaystyle \norm{\mathbf{q}_i^{\text{u}}-\mathbf{p}_{\text{e}}} \leq \frac{\Vmax^2}{2\Amax}, \\[10pt]
    \displaystyle\frac{\norm{\mathbf{q}_i^{\text{u}}-\mathbf{p}_{\text{e}}}}{\Vmax} + \frac{\Vmax}{2\Amax}\text{ otherwise}
    \end{cases}
\end{equation}
As the matrix $\mathbf{X}$ equivalently represents a bounding sequence $\hat{S}$, we formulate the MILP cost function $\hat{C}$ in~\cref{eq:hatcfull} as:
\begin{equation}\label{eq:hatcfull}
    \hat{C}(\mathbf{X}) = \frac{1}{2}\sum_{i=0}^{n^{\text{u}}-1}\!\left(\sum_{j=0}^{n^{\text{u}}}\!\left(c_{0,i}\,x_{i,0,j}+c_{i,0}\, x_{i,j,0}+\sum_{k=0}^{n^{\text{u}}} c_{i,j,k}\,x_{i,j,k}\right)\right).
\end{equation}
Such a cost function is a valid lower-bound on $\hat{C}_{\text{rel}}$, as the travel cost incurred for each consecutive pair of targets in $\hat{S}$ is bounded by an average value of two elements of $\mathbf{C}$, which themselves are lower-bounds on time-optimal travel costs. 

\begin{figure}
\begin{align}
  R_{\text{UB}}(S) = \max_{x \in \mathbf{X}} &\sum_{i=0}^{n^{\text{u}}-1} \left( r^{\text{u}}_i \sum_{j=0}^{n^{\text{u}}} \sum_{k=0}^{n^{\text{u}}} x_{i,j,k}\right)&&  \label{mip:eq1} \\
\text{s.t. }  & \hat{C}(\mathbf{X}) \leq \,\hat{T}_{\max}(S) \label{mip:eq2}\\
                    &  \sum_{i=0}^{n^{\text{u}}-1}\sum_{k=0}^{n^{\text{u}}}x_{i,0,k} = 1\label{mip:eq3}\\
                    &  \sum_{i=0}^{n^{\text{u}}-1}\sum_{j=0}^{n^{\text{u}}} x_{i,j,0}  = 1 \label{mip:eq4}\\
                    &\sum_{j=0}^{n^{\text{u}}}\sum_{k=0}^{n^{\text{u}}} x_{i,j,k} \leq 1 ~\forall i =0\ldots n^{\text{u}}-1\label{mip:eq5}\\
                    &\sum_{j=0}^{n^{\text{u}}-1}\sum_{k=0}^{n^{\text{u}}} x_{j,i+1,k} \leq 1 ~\forall i =0\ldots n^{\text{u}}-1\label{mip:eq6} \\
                    &\sum_{k=0}^{n^{\text{u}}-1}\sum_{j=0}^{n^{\text{u}}} x_{k,j,i+1} \leq 1 ~\forall i =0\ldots n^{\text{u}}-1  \label{mip:eq7} \\
                   & \sum_{j=0}^{n^{\text{u}}}\sum_{k=0}^{n^{\text{u}}} x_{i,j,k}= \sum_{j=0}^{n^{\text{u}}-1}\sum_{k=0}^{n^{\text{u}}} x_{j,i+1,k} ~\forall i = 0\ldots n^{\text{u}}-1 \label{mip:eq8}\\
                    & \sum_{j=0}^{n^{\text{u}}}\sum_{k=0}^{n^{\text{u}}} x_{i,j,k} = \sum_{k=0}^{n^{\text{u}}-1}\sum_{j=0}^{n^{\text{u}}} x_{k,j,i+1} ~\forall i = 0\ldots n^{\text{u}}-1 \label{mip:eq9}\\
                    & x_{i,j,k} \leq \sum_{l=0}^{n^{\text{u}}} x_{j-1,l,i+1} ~\forall i = 0\ldots n^{\text{u}}-1~\forall j,k = 1\ldots n^{\text{u}} \label{mip:eq10}\\
                    & x_{i,j,k} \leq \sum_{l=0}^{n^{\text{u}}} x_{k-1,i+1,l}~\forall i = 0\ldots n^{\text{u}}-1~\forall j,k = 1\ldots n^{\text{u}}\label{mip:eq11}
\end{align}
\end{figure}

The upper bound $R_{\text{UB}}(S)$ is obtained by solving the MILP in~\cref{mip:eq1,mip:eq2,mip:eq3,mip:eq4,mip:eq5,mip:eq6,mip:eq7,mip:eq8,mip:eq9,mip:eq10,mip:eq11}.
\Cref{mip:eq1} maximizes the reward collected by the primitives selected by $\mathbf{X}$; \cref{mip:eq2} constraints the travel budget $\hat{T}_{\max}(S)$, \Cref{mip:eq3,mip:eq4} enforce visiting $\mathbf{q}_{s_{\card{S}-1}}$ and $\mathbf{p}_{\text{e}}$, and \cref{mip:eq5,mip:eq6,mip:eq7,mip:eq8,mip:eq9} collect each reward $r_i^{\text{u}}$ at most once via a continuous target sequence.
\Cref{mip:eq10,mip:eq11} enforce the trajectory primitive overlap.

Given the second-order motion dynamics, a high-quality solution is expected to maximize velocity and minimize changes in direction.
Since isolated sub-tours require the vehicle to turn, high-quality MILP solutions are expected to naturally prevent their occurrence.
Therefore, we implement lazy constraints for sub-tour elimination, avoiding the combinatorial cost of including them exhaustively.

\subsection{NLP formulations of the time-optimal control problems}\label{sec:se:nlp}
The BnB procedure relies on time-optimal trajectories through targets with varying start and end constraints.
For the discrete model~\eqref{eq:model:ade}, the time-optimal trajectory between two targets satisfying Pontryagin's minimum principle is bang-zero-bang~\citep{hehn2012performance}, the representation behind current state-of-the-art time-optimal quad-rotor planning~\citep{teissing2024real}.
For many targets,~\cite{qin2024time} showed that a minimum number of input-switching sub-steps $\nsub$ between each pair is required for time-optimality, proportional to the system order and constraints.

To apply this input-switching trajectory, we add a per-step duration input $\Delta t_k \in \mathbb{R}_{\geq 0}$ in~\cref{eq:state_propagation}:
\begin{subequations}
\label{eq:state_propagation}
\begin{align}
  \mathbf{p}_{k+1} &= \mathbf{p}_{k}+ \Delta t_k\,\mathbf{v}_k + \frac{\Delta t_k^2}{2}\, \mathbf{a}_k,\\
  \mathbf{v}_{k+1} &= \mathbf{v}_{k}+ \Delta t_k\,\mathbf{a}_k,
\end{align}
\end{subequations}
where $\Delta t_k$ is the variable per-step duration (generalizing the fixed $\Delta t$ of~\cref{eq:model:ade}), and the total number of steps is $\nsub$ times the number of visited positions.

To choose $\nsub$, we follow~\cite{hehn2012performance}, who show that the minimal quadrotor with collective thrust requires at most 5 switches for a time-optimal fixed-end-point trajectory.
Although our model~\eqref{eq:state_propagation} neglects rotation, the velocity constraint~\eqref{eq:velmag} can still cause singular input arcs in a time-optimal trajectory (requirement for infinitely many input switches), e.g.\ when following an arc of fixed radius.
In our experiments, increasing the switches beyond 5 did not improve quality, so we set $\nsub=5$ empirically.

Unlike the heuristic representation~\eqref{sec:heuristic_trajectory}, the optimizations based on~\cref{eq:state_propagation} solve all axes simultaneously, improving quality and flexibility in relaxed costs at the expense of greater complexity and reliance on off-the-shelf solvers.
The sequence cost~\eqref{eq:statement:cost_function} is implemented as the NLP
\begin{equation}\label{eq:nlp:c_full}
   \begin{array}{lll}
      \displaystyle C\textsuperscript{NLP}(S) = & \multicolumn{2}{l}{\displaystyle \min_{\mathbf{p}, \mathbf{v}, \mathbf{a}, \Delta t}\sum_{k=0}^{\nsub(\card{S}+1) - 1}\Delta t_k}\\
      &\text{s.t.} & \mathbf{p}_0 = \mathbf{p}_{\text{s}},~\mathbf{p}_{\nsub(\card{S}+1)} = \mathbf{p}_{\text{e}}, \\
      && \mathbf{v}_0 = \mathbf{0},~\mathbf{v}_{\nsub(\card{S}+1)} = \mathbf{0},\\
      && {\mathbf{p}_{\nsub (i+1)} = \mathbf{q}_{s_{i}} }\forall i=0\ldots\card{S}-1,\\
      && \text{and~\cref{eq:constraints,eq:state_propagation}.}
   \end{array}
\end{equation}
The end-velocity relaxed cost~\eqref{eq:se:crel} becomes
\begin{equation}\label{eq:nlp:c_rel}
   \begin{array}{lll}
      \displaystyle C_{\text{rel}}^{\text{NLP}}(S) =& \multicolumn{2}{l}{\displaystyle \min_{\mathbf{p}, \mathbf{v}, \mathbf{a}, \Delta t} \sum_{k=0}^{\nsub\card{S} - 1} \Delta t_k}\\
      &\text{s.t.} & \mathbf{p}_0 = \mathbf{p}_{\text{s}},~\mathbf{v}_0 = \mathbf{0},\\
      &&  {\mathbf{p}_{\nsub (i+1)} = \mathbf{q}_{s_{i}} }\forall i=0\ldots\card{S}-1,\\
      && \text{and~\cref{eq:constraints,eq:state_propagation}.}
   \end{array}
\end{equation}
The start-velocity relaxed cost of a bounding sequence $\hat{S}$~\eqref{eq:se:crelhat} becomes
\begin{equation}\label{eq:nlp:c_hat_rel}
   \begin{array}{lll}
      \displaystyle\hat{C}_{\text{rel}}^{\text{NLP}}(\hat{S}) =&  \multicolumn{2}{l}{\displaystyle \min_{\mathbf{p}, \mathbf{v}, \mathbf{a}, \Delta t} \sum_{k=0}^{\nsub\card{\hat{S}} - 1} \Delta t_k}\\
      &\text{s.t.} & \mathbf{p}_0=\mathbf{q}_{\hat{s}_0},~\mathbf{p}_{\nsub\card{\hat{S}}} = \mathbf{p}_{\text{e}}, \\
      && \mathbf{v}_{\nsub\card{\hat{S}}} = \mathbf{0},\\
      && \mathbf{p}_{\nsub i} = \mathbf{q}_{\hat{s}_{i}}\forall i=1\ldots\card{\hat{S}}-1,\\
      && \text{and~\cref{eq:constraints,eq:state_propagation}.}
   \end{array}
\end{equation}
The primitive cost~\eqref{eq:se:cprimitive} is implemented as~\cref{eq:nlp:c_primitive}.
Since the primitives are pre-computed before BnB, we generalize the implementation to arbitrary positions $\mathbf{q}_{\text{A}}$, $\mathbf{q}_{\text{B}}$, $\mathbf{q}_{\text{C}}$ corresponding to the indices $j$, $i$, $k$:
\begin{equation}\label{eq:nlp:c_primitive}
   \begin{array}{llll}
      \displaystyle c^{\text{NLP}}(\mathbf{q}_{\text{A}},\mathbf{q}_{\text{B}},\mathbf{q}_{\text{C}})=&\multicolumn{2}{l}{\displaystyle\min_{\mathbf{p}, \mathbf{v}, \mathbf{a}, \Delta t}  \sum_{m=0}^{2\nsub - 1} \Delta t_{m}} \\
      &\text{s.t.} & \mathbf{p}_0 &= \mathbf{q}_{\text{A}},\\
      &&\mathbf{p}_{\nsub} &= \mathbf{q}_{\text{B}},\\
      &&\mathbf{p}_{2\nsub} &= \mathbf{q}_{\text{C}},\\
      &&\mathbf{v}_{2\nsub} &= \mathbf{0} \text{ if } \mathbf{q}_{\text{C}} = \mathbf{p}_{\text{e}},\\
      && \multicolumn{2}{l}{\text{and~\cref{eq:constraints,eq:state_propagation}.}}
   \end{array}
\end{equation}

\section{Results}\label{sec:results}
This section presents the computational evaluation of the solvers and data from a real-world multi-rotor UAV deployment.
\Cref{sec:results:kop} compares the solvers with the KOP state of the art; \cref{sec:results:ce} evaluates them on random DVOP instances over travel budget, maximal acceleration, and gravity; \cref{sec:results:bnb} discusses the BnB complexity and the MILP lower-bound quality; \cref{sec:results:lns} reports large-instance LNS solutions; and \cref{sec:results:real} validates the trajectory on a real UAV.

All computational results were obtained on a PC with an Intel i7-13700KF CPU and 64 GB of RAM. The MILP~(\cref{sec:se:mip}) was implemented with the Gurobi 13.0 optimizer~\citep{gurobi}, and the NLPs~(\Cref{sec:se:nlp}) with IPOPT~\citep{ipopt} and the HSL MA97 linear solver~\citep{hsl_collection}. Both solvers are implemented in C++ and provided with benchmark datasets\footnote{\url{https://github.com/fnekovar/DVOP_code_repository}}.

\subsection{Kinematic Orienteering Problem}\label{sec:results:kop}
The original KOP~\citep{meyer2022kinematic} seeks a kinematic trajectory maximizing the sum of location priorities (rewards) under a maximum flight time, selecting from a set of velocity vectors per location and using axis synchronization for feasible trajectories and times.
In their evaluation, per-axis velocity and acceleration constraints were scaled by $\frac{1}{\sqrt{2}}$ for fair comparison to the DOP, as maximal acceleration over both axes could breach the DOP acceleration magnitude constraint.
For multi-rotor UAVs, however, acceleration is mainly a function of thrust, so the magnitude constraints of~\cite{nekovavr2023multi,ghotavadekar2024variable} are more suitable, representing the maximum thrust as the maximum acceleration magnitude.

For a complete comparison, we solve the KOP under both constraint formulations on the Tsiligirides Set 2 dataset~\citep{tsiligirides1984dataset}.
Per-axis constraints ($v_{\text{max}}^i=\frac{3}{\sqrt{2}}\si{\meter\per\second}, a_{\text{max}}^i=\frac{1.5}{\sqrt{2}}\si{\meter\per\second\squared},\forall i\in\{x,y\}$) are in~\cref{tbl:kop:pa}, and magnitude constraints ($\Vmax=3\si{\meter\per\second}, \Amax=1.5\si{\meter\per\second}$) in~\cref{tbl:kop:magnitude}, where $R$ is the maximal reward from the solver given in the subscript and $t$ the computation time.
KOP-1 and KOP-6 are the best solvers of~\cite{meyer2022kinematic}: the exact KOP solver limited to maximal velocity (KOP-1) and an LNS heuristic limited to 6 traversal velocities (KOP-6).
VT-MPC is the variable time-step solver of~\cite{ghotavadekar2024variable}.
The KOP-6 and proposed LNS solvers were run 10 times per instance, with the average reward shown in brackets.

\begin{table*}[!htb] 
    \centering
    \normalsize
    {\renewcommand{\tabcolsep}{5.8pt} 
    \renewcommand{\arraystretch}{0.75}
    \caption{KOP performance benchmark with per-axis acceleration constraints\vspace{-1em}} 
    \label{tbl:kop:pa}
    \begin{tabular}{crrrrrr}
    \toprule 
    \Cmax [\si{\second}]& \multicolumn{1}{c}{$R_{\text{KOP-1}}$} & \multicolumn{1}{c}{$R_{\text{KOP-6}}$} & \multicolumn{1}{c}{$R_{\text{LNS}}$} & \multicolumn{1}{c}{$R_{\text{BnB}}$} & \multicolumn{1}{c}{$t\textsubscript{LNS}$ [\si{s}]} & \multicolumn{1}{c}{$t\textsubscript{BnB}$ [\si{h}]} \\
    \midrule
    10 & 95 & 80 (75) & 125 (122.5) & \textbf{130} & 0.45 & 0.08 \\
    15 & 180 & 165 (165) & 220 (217.5) & \textbf{230} & 0.61 & 0.68 \\
    20 & 250 & 250 (237.5) & 325 (321.5) & \textbf{340} & 0.92 & 1.54 \\
    25 & 325 & 330 (316.5) & 380 (374.5) & \textbf{390} & 1.02 & 18.41\\
    30 & 390 & 390 (377.5) & 425 (417) & \textbf{440} & 1.18 & 44.24\\
    35 & 430 & 435 (422.5) & \textbf{450 (450)} & \textbf{450} & 1.11 & 0.15 \\
    40 & \textbf{450} & \textbf{450 (450)} & \textbf{450 (450)} & \textbf{450} & 0.96 & 1.68 \\
    
    \bottomrule 
    \end{tabular} 
    } 
    \vspace{-1.0em}
\end{table*}
\begin{table*}[!htb] 
    \centering
    \normalsize
    {\renewcommand{\tabcolsep}{5.8pt} 
    \renewcommand{\arraystretch}{0.75}
    \caption{KOP performance benchmark with acceleration magnitude constraints\vspace{-1em}} 
    \label{tbl:kop:magnitude}
    \begin{tabular}{crrrrr}
    \toprule 
    \Cmax [\si{s}] & \multicolumn{1}{c}{$R_{\text{VT-MPC}}$} & \multicolumn{1}{c}{$R_{\text{LNS}}$} & \multicolumn{1}{c}{$R_{\text{BnB}}$} & \multicolumn{1}{c}{$t\textsubscript{LNS}$ [\si{s}]} & \multicolumn{1}{c}{$t\textsubscript{BnB}$ [\si{\hour}]} \\
    \midrule
    10 &  115 &  135 (133) & \textbf{150} & 2.05 & 0.38\\
    15 &  215 &  270 (256.5) & \textbf{295} & 2.47 & 0.82\\
    20 &  345 & \textbf{355} (343) & \textbf{355} & 3.32 & 24.80\\
    25 &  395 &  400 (392) & \textbf{430} & 3.44 & 81.02\\
    30 &  430 &  \textbf{450} (446) & \textbf{450} & 3.86 & 34.82\\
    35 &  \textbf{450} & \textbf{450 (450)} & \textbf{450} & 3.27 &0.86\\
    40 &  \textbf{450} & \textbf{450 (450)} & \textbf{450} & 3.33 & 18.96\\
    \bottomrule 
    \end{tabular} 
    } 
\end{table*}

Both proposed solvers improve solution quality.
The LNS exceeds the state of the art in a few seconds, while the BnB yields the highest quality overall at the cost of up to tens of hours.

\subsection{Computational comparison of the proposed solvers}\label{sec:results:ce}
We evaluate the solvers on $n_{\text{inst}}=20$ random 3D instances.
The start and end are fixed at the origin, $\mathbf{p}_{\text{s}}=\mathbf{p}_{\text{e}}=\mathbf{0}$.
To avoid bias toward any spatial pattern (as in \cref{sec:results:kop}), we sample 15 targets uniformly within a sphere of radius \SI{10}{\meter} around the origin, and assign each a reward of 1.
We set $\Vmax = \SI{3}{\meter\per\second}$ throughout.
For each combination of \Cmax~and \Amax, results are averaged over all $n_{\text{inst}}$ instances in~\cref{tbl:lns_bnb_comparison}.

\begin{table*}[!h]
    \centering
    \caption{Computational evaluation on 20 randomized instances.}
    \subfloat[Without gravity acceleration\label{tbl:lns_bnb_comparison_nograv}]{%
        \begin{minipage}{0.49\textwidth}
            \centering
                {\renewcommand{\tabcolsep}{4pt}%
\renewcommand{\arraystretch}{0.8}%
\begin{tabular}{ccrrrrr}
\toprule
\Amax & \Cmax  & \multicolumn{1}{c}{$R^{\text{avg}}_{\text{LNS}}$} & \multicolumn{1}{c}{$R^{\text{avg}}_{\text{BnB}}$} & \multicolumn{1}{c}{$PDB$} & \multicolumn{1}{c}{$t^{\text{avg}}_{\text{LNS}}$} & \multicolumn{1}{c}{$t^{\text{avg}}_{\text{BnB}}$}\\\relax
[\si{\meter\per\second\squared}] & [\si{\second}] & \multicolumn{1}{c}{[-]} & \multicolumn{1}{c}{[-]} & \multicolumn{1}{c}{[\%]} & \multicolumn{1}{c}{[\si{\second}]} & \multicolumn{1}{c}{[\si{\second}]}\\
\midrule
1.5 & 10 & 2.00 & \textbf{2.30} & 11.7 & 0.2 & 3.6\\
1.5 & 15 & 4.45 & \textbf{4.85} & 7.7 & 1.1 & 37.7\\
1.5 & 20 & 6.60 & \textbf{7.15} & 7.4 & 1.9 & 232.0\\
1.5 & 25 & 8.70 & \textbf{9.25} & 5.7 & 2.5 & 702.9\\
1.5 & 30 & 10.20 & \textbf{11.15} & 8.5 & 3.2 & 1413.2\\
1.5 & 35 & 11.95 & \textbf{13.00} & 8.1 & 3.4 & 1691.6\\
1.5 & 40 & 13.15 & \textbf{14.25} & 7.7 & 2.9 & 4730.9\\
\midrule
3.0 & 10 & 3.40 & \textbf{3.90} & 12.3 & 0.2 & 8.6\\
3.0 & 15 & 6.20 & \textbf{6.90} & 9.9 & 1.0 & 69.0\\
3.0 & 20 & 8.70 & \textbf{9.50} & 8.4 & 1.6 & 204.7\\
3.0 & 25 & 10.85 & \textbf{11.85} & 8.5 & 2.2 & 278.8\\
3.0 & 30 & 12.75 & \textbf{13.55} & 5.8 & 2.0 & 944.1\\
3.0 & 35 & 14.05 & \textbf{14.75} & 4.7 & 1.8 & 2159.1\\
3.0 & 40 & \textbf{15.00} & \textbf{15.00} & 0.0 & 1.6 & 9.7\\
\midrule
4.5 & 10 & 4.00 & \textbf{4.50} & 10.5 & 0.2 & 12.4\\
4.5 & 15 & 6.90 & \textbf{7.55} & 8.8 & 1.0 & 67.2\\
4.5 & 20 & 9.30 & \textbf{10.05} & 7.4 & 1.5 & 160.5\\
4.5 & 25 & 11.55 & \textbf{12.40} & 6.8 & 1.7 & 381.6\\
4.5 & 30 & 13.35 & \textbf{14.00} & 4.7 & 1.9 & 520.5\\
4.5 & 35 & 14.70 & \textbf{15.00} & 2.0 & 2.1 & 232.0\\
4.5 & 40 & \textbf{15.00} & \textbf{15.00} & 0.0 & 1.1 & 8.9\\
\midrule
6.0 & 10 & 4.05 & \textbf{4.80} & 15.1 & 0.2 & 11.5\\
6.0 & 15 & 7.10 & \textbf{7.95} & 10.7 & 0.9 & 69.3\\
6.0 & 20 & 9.85 & \textbf{10.45} & 5.6 & 1.6 & 197.2\\
6.0 & 25 & 11.90 & \textbf{12.85} & 7.4 & 1.9 & 204.9\\
6.0 & 30 & 13.80 & \textbf{14.40} & 4.1 & 1.9 & 447.2\\
6.0 & 35 & 14.90 & \textbf{15.00} & 0.7 & 1.6 & 102.7\\
6.0 & 40 & \textbf{15.00} & \textbf{15.00} & 0.0 & 1.0 & 10.1\\
\bottomrule
\end{tabular}%
}

        \end{minipage}
    }
    \subfloat[With gravity acceleration\label{tbl:lns_bnb_comparison_grav}]{%
        \begin{minipage}{0.49\textwidth}
            \centering
                {
\renewcommand{\tabcolsep}{4pt}%
\renewcommand{\arraystretch}{0.8}%
\begin{tabular}{ccrrrrr}
\toprule
\Amax & \Cmax  & \multicolumn{1}{c}{$R^{\text{avg}}_{\text{LNS}}$} & \multicolumn{1}{c}{$R^{\text{avg}}_{\text{BnB}}$} & \multicolumn{1}{c}{$PDB$} & \multicolumn{1}{c}{$t^{\text{avg}}_{\text{LNS}}$} & \multicolumn{1}{c}{$t^{\text{avg}}_{\text{BnB}}$}\\\relax
[\si{\meter\per\second\squared}] & [\si{\second}] & \multicolumn{1}{c}{[-]} & \multicolumn{1}{c}{[-]} & \multicolumn{1}{c}{[\%]} & \multicolumn{1}{c}{[\si{\second}]} & \multicolumn{1}{c}{[\si{\second}]}\\
\midrule
1.5+g & 10 & 3.70 & \textbf{4.50} & 17.3 & 0.4 & 21.7\\
1.5+g & 15 & 6.55 & \textbf{7.55} & 13.0 & 1.6 & 84.6\\
1.5+g & 20 & 8.90 & \textbf{10.05} & 11.4 & 2.5 & 192.1\\
1.5+g & 25 & 11.15 & \textbf{12.35} & 9.7 & 3.0 & 446.5\\
1.5+g & 30 & 12.95 & \textbf{14.05} & 7.8 & 3.1 & 572.8\\
1.5+g & 35 & 14.40 & \textbf{15.00} & 4.0 & 2.9 & 482.4\\
1.5+g & 40 & \textbf{15.00} & \textbf{15.00} & 0.0 & 2.4 & 10.8\\
\midrule
3.0+g & 10 & 4.10 & \textbf{4.90} & 16.7 & 0.3 & 14.9\\
3.0+g & 15 & 7.15 & \textbf{8.10} & 11.8 & 1.3 & 88.0\\
3.0+g & 20 & 9.75 & \textbf{10.70} & 8.8 &  2.1 & 194.6\\
3.0+g & 25 & 11.90 & \textbf{12.90} & 7.8 & 2.6 & 186.8\\
3.0+g & 30 & 13.80 & \textbf{14.60} & 5.5 & 2.8 & 307.8\\
3.0+g & 35 & 14.85 & \textbf{15.00} & 1.0 & 2.2 & 103.4\\
3.0+g & 40 & \textbf{15.00} & \textbf{15.00} & 0.0 & 1.4 & 9.2\\
\midrule
4.5+g & 10 & 4.25 & \textbf{4.95} & 14.0 & 0.3 & 19.2\\
4.5+g & 15 & 7.35 & \textbf{8.25} & 10.7 & 1.1 & 73.3\\
4.5+g & 20 & 9.95 & \textbf{10.85} & 8.2 & 1.8 & 152.3\\
4.5+g & 25 & 12.10 & \textbf{12.95} & 6.6 & 2.1 & 278.7\\
4.5+g & 30 & 13.95 & \textbf{14.60} & 4.4 & 2.2 & 356.2\\
4.5+g & 35 & 14.95 & \textbf{15.00} & 0.3 & 1.9 & 38.4\\
4.5+g & 40 & \textbf{15.00} & \textbf{15.00} & 0.0 & 1.2 & 12.5\\
\midrule
6.0+g & 10 & 4.30 & \textbf{5.00} & 13.5 & 0.3 & 22.0\\
6.0+g & 15 & 7.55 & \textbf{8.30} & 9.0 & 1.2 & 71.8\\
6.0+g & 20 & 10.05 & \textbf{10.90} & 7.8 & 1.8 & 130.8\\
6.0+g & 25 & 12.10 & \textbf{13.00} & 6.9 & 2.0 & 298.6\\
6.0+g & 30 & 14.00 & \textbf{14.70} & 4.8 & 2.0 & 288.3\\
6.0+g & 35 & 14.95 & \textbf{15.00} & 0.3 & 1.7 & 108.5\\
6.0+g & 40 & \textbf{15.00} & \textbf{15.00} & 0.0 & 1.1 & 9.6\\
\bottomrule
\end{tabular}%
}

        \end{minipage}
    }
    \label{tbl:lns_bnb_comparison}
    \vspace{-2.0em}
\end{table*}
\Cref{tbl:lns_bnb_comparison_nograv} gives results without gravity, and~\cref{tbl:lns_bnb_comparison_grav} with gravity $g = \SI{10}{\meter\per\square\second}$.
The solvers' average rewards are $R^{\text{avg}}_{\text{LNS}}$ and $R^{\text{avg}}_{\text{BnB}}$ over all instances with given \Amax~and \Cmax.
With per-instance rewards $R^*_{\text{LNS},i}$ and $R^*_{\text{BnB},i}$, the average percentage deviation of LNS from BnB is $PDB = \sum_{i} PDB_i \div n_{\text{inst}}$, $PDB_i = 100 \times (R^*_{\text{BnB},i} - R^*_{\text{LNS},i}) \div R^*_{\text{BnB},i}$.
Average computation times are $t^{\text{avg}}_{\text{LNS}}$ and $t^{\text{avg}}_{\text{BnB}}$.
The LNS rewards initialize the BnB lower bound $R_{\text{LB}}$; the LNS was run 10 times per instance and the best result used.

\begin{figure*}[!h]
    \subfloat[BnB solver]{%
        \includegraphics[width=.5\linewidth]{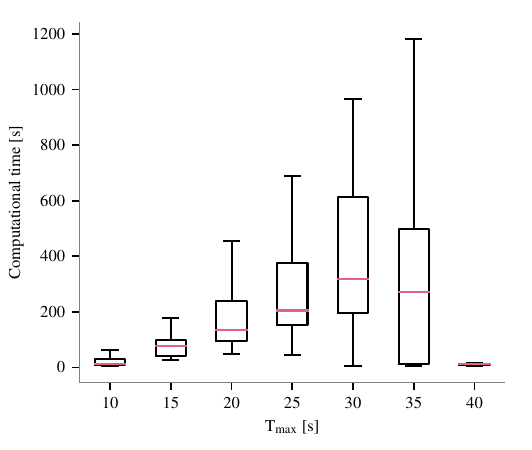}%
        \label{subfig:amax:a}%
    }
    \subfloat[LNS solver]{%
        \includegraphics[width=.5\linewidth]{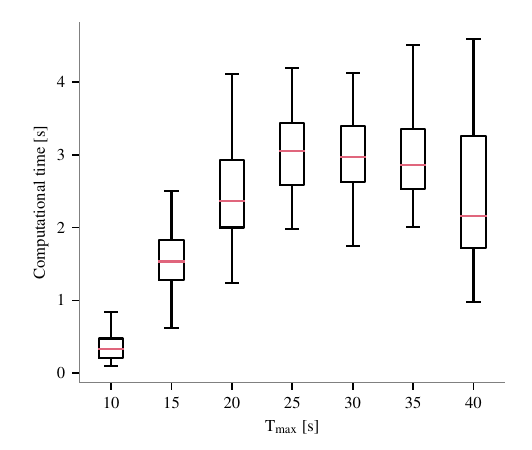}%
        \label{subfig:amax:b}%
    }
    \caption{Box-plots of the BnB~\protect\subref{subfig:amax:a} and LNS~\protect\subref{subfig:amax:b} computational times over all randomized instances with gravity (presented in~\cref{tbl:lns_bnb_comparison_grav}) with regard to the travel budget \Cmax~and acceleration magnitude constraint $\Amax=g+\SI{1.5}{\meter\per\square\second}$.
    }
    \label{fig:results:box_amax}
\end{figure*}

For all configurations, the LNS PDB is $17.3\%$ for the tightest instance and decreases as \Cmax~and/or \Amax~increase.
The BnB computation time is generally three orders of magnitude higher than the LNS.
Neither solver's time scales strictly with \Cmax; instead, it peaks as the collected reward approaches the total $\sum R = \sum_{i \in I} r_i$ (\cref{fig:results:box_amax}), a trend also seen in the KOP results~(\cref{tbl:kop:pa,tbl:kop:magnitude}).

\subsection{Branch-and-Bound evaluation}\label{sec:results:bnb}

To evaluate the BnB, we consider two gaps: the MILP cost-function gap (from the trajectory primitives) and the reward upper-bound gap.
They are related but distinct: the cost gap is isolated to the MILP, while the reward gap also depends on the relaxation of the first-order continuity constraint at the last target of a sequence.

\begin{figure*}[h]
\centering
    \subfloat[]{%
        \includegraphics[width=.5\linewidth]{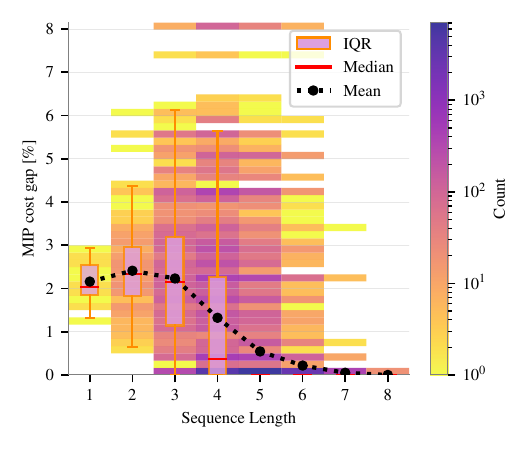}%
        \label{subfig:mipcost_gap_len}%
    }
    \subfloat[]{%
        \includegraphics[width=.5\linewidth]{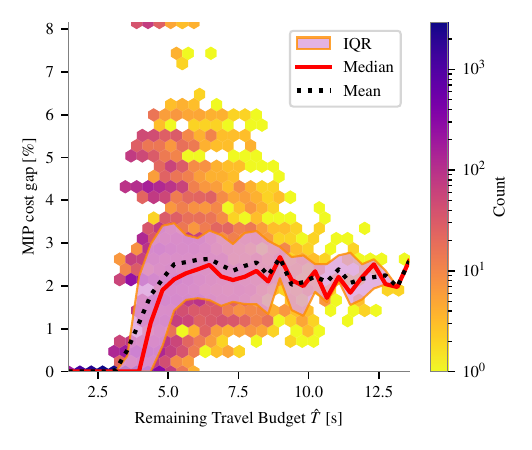}%
        \label{subfig:mipcost_gap_budget}%
    }
    \caption{Visualization of MILP cost function gap distribution with regard to the sequence length~\protect\subref{subfig:mipcost_gap_len} and the MILP travel budget~\protect\subref{subfig:mipcost_gap_budget} for one randomized instance with gravity. The instance parameters are: $\Cmax=\SI{15}{s}$ and $\Amax=1.5+g\,\si{\meter\per\square\second}$.}
    \label{fig:results:mipcost}
    \vspace{-1em}
\end{figure*}
\Cref{fig:results:mipcost} shows the distribution of the MILP gap over all valid sequences from exhaustive exploration of one instance, reporting mean, median, and \emph{inter-quartile range} (IQR).
\Cref{subfig:mipcost_gap_len} plots the cost gap $\delta_{\hat{C}} = 100 \times (\hat{C}_{\text{rel}}-\hat{C}) \div \hat{C}_{\text{rel}}$ against the sequence length $|S|$, and \cref{subfig:mipcost_gap_budget} against the budget $\hat{T}_{\max}(S)$.
Since \cref{subfig:mipcost_gap_budget} shows $\delta_{\hat{C}}$ plateauing, we report the per-instance cost gap as the mean $\bar{\delta}_{\hat{C}}$ over all sequences.
Likewise, the mean reward upper-bound gap is $\bar{\delta}_{R_{\text{UB}}}$ from $\delta_{R_{\text{UB}}} = 100 \times (R_{\text{UB}}-R_{\text{UB}}^*) \div R_{\text{UB}}^*$, where $R_{\text{UB}}^*(S)$ is the true upper bound from exhaustive search (the best reward achievable by further branching).

\begin{table*}[h]
    \centering
    \caption{Computational evaluation of the Branch-and-Bound solver quality}
    {\renewcommand{\tabcolsep}{5.8pt}%
\renewcommand{\arraystretch}{0.8}%
\begin{tabular}{ccrrrrrrrr}
\toprule
\Amax & \Cmax & \multicolumn{1}{c}{$\bar{\delta}_{R_{\text{UB}}}$} & \multicolumn{1}{c}{$\max\delta_{R_{\text{UB}}}$} & \multicolumn{1}{c}{$\bar{\delta}_{\hat{C}}$} & \multicolumn{1}{c}{$\max\delta_{\hat{C}}$} & \multicolumn{1}{c}{$N_{\text{open}}^{\text{exh}}$} & \multicolumn{1}{c}{$N^{\text{BnB}}_{\text{open}}$} & \multicolumn{1}{c}{$N_{\text{cut}}$} \\\relax
[\si{\meter\per\second\squared}] & [\si{\second}] & \multicolumn{1}{c}{[\%]} & \multicolumn{1}{c}{[\%]} & \multicolumn{1}{c}{[\%]} & \multicolumn{1}{c}{[\%]} & \multicolumn{1}{c}{[-]} & \multicolumn{1}{c}{[-]} & \multicolumn{1}{c}{[\%]} \\
\midrule
1.5+g & 10 & 7.09 & 50.00 & 1.98 & 15.16 & 3733 & 65 & 98.26 \\ 
1.5+g & 15 & 4.27 & 66.67 & 2.00 & 15.16 & 141483 & 210 & 99.85 \\ 
1.5+g & 20 & 2.99 & 50.00 & 2.01 & 16.76 & 3899041 & 180 & 99.99 \\ 
3.0+g & 10 & 3.51 & 50.00 & 0.91 & 8.15 & 5833 & 129 & 97.79 \\ 
3.0+g & 15 & 2.15 & 50.00 & 0.94 & 8.15 & 250381 & 131 & 99.95 \\ 
4.5+g & 10 & 1.66 & 50.00 & 0.56 & 5.53 & 7135 & 65 & 99.09 \\ 
4.5+g & 15 & 1.59 & 33.33 & 0.63 & 7.54 & 307179 & 131 & 99.96 \\ 
6.0+g & 10 & 1.95 & 50.00 & 0.43 & 4.20 & 7746 & 79 & 98.98 \\ 
6.0+g & 15 & 1.24 & 33.33 & 0.48 & 4.53 & 346076 & 104 & 99.97 \\ 
\bottomrule
\end{tabular}
}
    \label{tbl:bnb_evaluation}
\end{table*}
Gap values across instances are in~\cref{tbl:bnb_evaluation}; $\max\delta_{\hat{C}}$ and $\max\delta_{R_{\text{UB}}}$ are the maxima.
$N_{\text{open}}^{\text{exh}}$ and $N_{\text{open}}^{\text{BnB}}$ count nodes opened in exhaustive and BnB exploration, and $N_{\text{cut}} = 100 \times (N_{\text{open}}^{\text{exh}}-N_{\text{open}}^{\text{BnB}}) \div N_{\text{open}}^{\text{exh}}$ is the percentage of nodes cut by the upper bound.

Both $\bar{\delta}_{\hat{C}}$ and $\bar{\delta}_{R_{\text{UB}}}$ decrease as \Amax~increases, confirming that relaxing the acceleration constraint reduces the DVOP to the OP.
As \Cmax~increases, the reward gap decreases while the MILP cost gap increases.
The cut fraction $N_{\text{cut}}$ exceeds \SI{97}{\percent} in all instances and grows with both \Amax~and \Cmax, so the upper bound tightens with the solution-space size.

\subsection{Large Neighborhood Search evaluation}\label{sec:results:lns}
\begin{figure*}[ht!]
    \centering
    \subfloat[Tsiligirides Set 2~\cite{tsiligirides1984dataset}]{%
        \includegraphics[width=.5\linewidth]{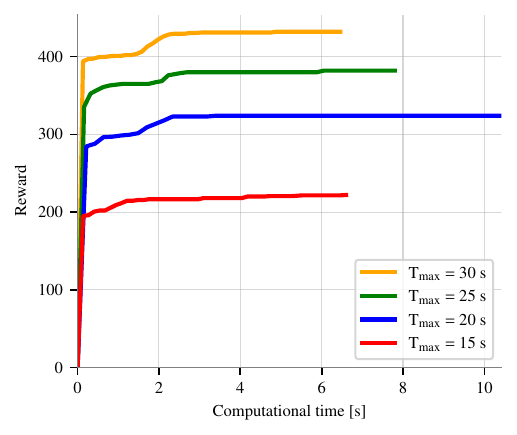}
        \label{fig:reward_progress:ts2}
    }
    \subfloat[Chao Set 64~\cite{chao1996dataset}]{%
        \includegraphics[width=.5\linewidth]{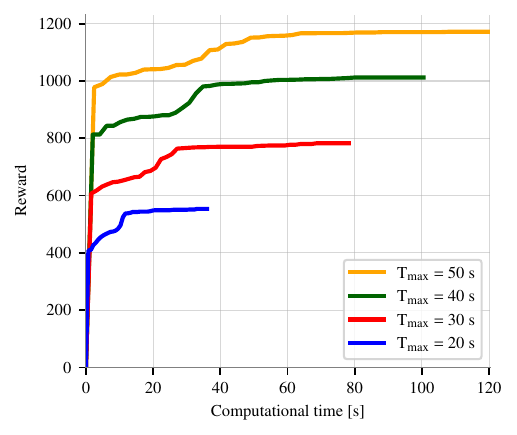}
        \label{fig:reward_progress:set64}
    }
    \caption{Reward progress for proposed anytime LNS heuristic solver.}
    \label{fig:reward_progress}
    \vspace{-1em}
\end{figure*}
\Cref{fig:reward_progress} shows the best reward found over time by the LNS ($\Vmax=\SI{3}{\meter\per\second}$) on Tsiligirides Set 2~\citep{tsiligirides1984dataset} (\cref{fig:reward_progress:ts2}) and Chao Set 64~\citep{chao1996dataset} (\cref{fig:reward_progress:set64}); each curve averages 10 runs.
The method improves steeply at first, reaching up to 90\% of the final reward within \SI{0.1}{\second}, then jumps again at the LNS-to-refinement transition (\cref{sec:lns,sec:refinement}), after which gains are marginal.
The LNS phase rapidly finds high-quality regions, while refinement extracts the remaining gains and fine-tunes the incumbent.

\subsection{Real-world trajectory validation}\label{sec:results:real}

\begin{figure}[h]
    \centering
    \includegraphics[width=0.5\linewidth]{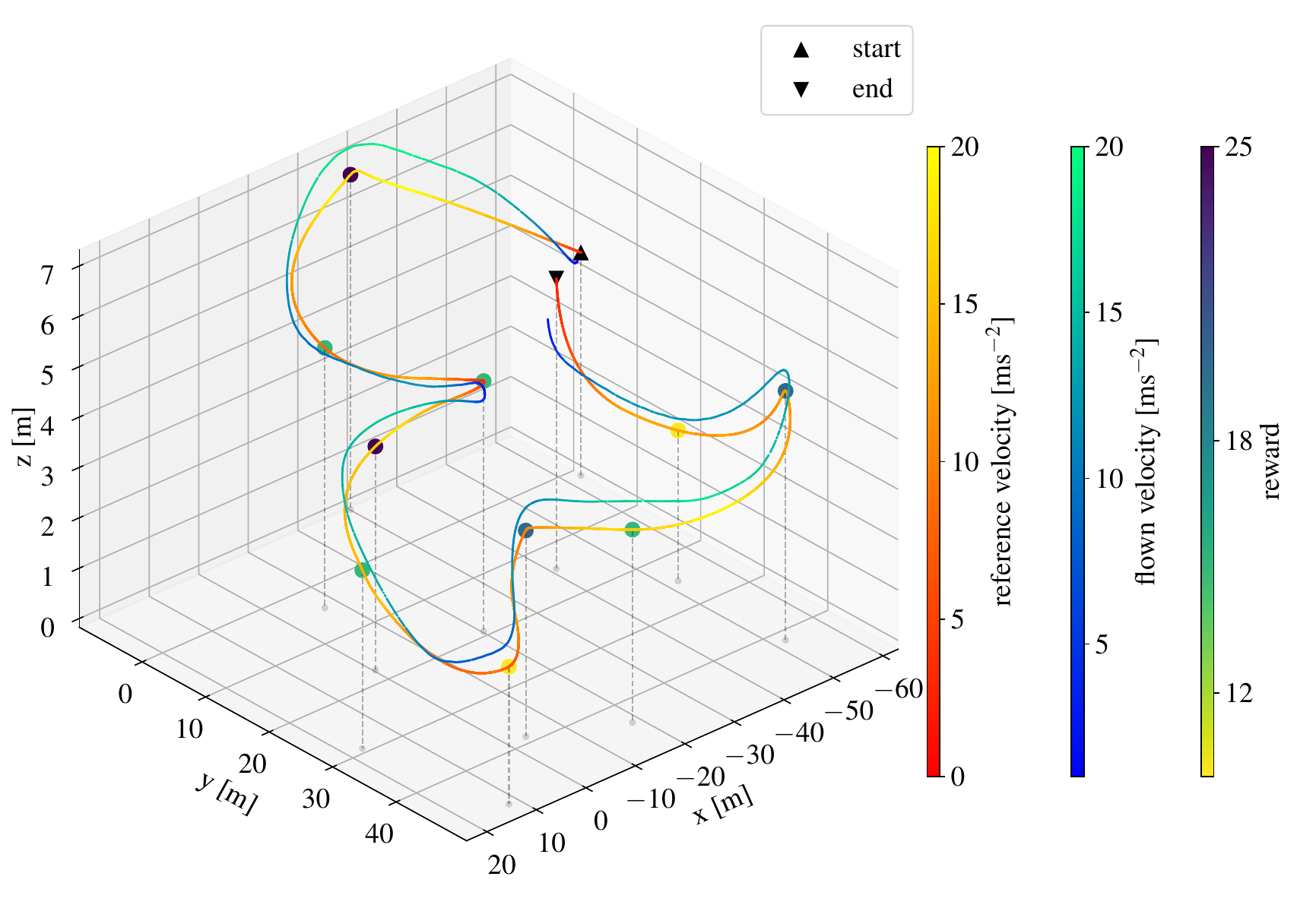}
    \caption{Visualization of a 3D problem solution flown in a real-world experiment. The warm-colored trajectory is the solver reference; the cold-colored trajectory was flown by the UAV.}
    \label{fig:rw:3d}
    \vspace{-1em}
\end{figure}

The point-mass trajectories from the solvers were validated in real flight on a custom dataset (\cref{fig:rw:3d}).
We used a \SI{330}{\milli\meter} quadcopter with a Khadas VIM3\footnote{https://www.khadas.com/vim3} onboard computer running the MRS UAV System~\citep{baca2021mrs, hert2023mrs}.
To assess the practical accuracy of the point-mass model, the UAV tracked pre-computed trajectories for different \Vmax, spanning conservative inspection to time-critical data collection.
\Cref{tbl:rw} reports the average root-mean-square tracking error (RMSE), i.e.\ the deviation from the planned trajectory due to model mismatch and disturbances such as wind gusts.
For inspection, we also report the average minimum distance to solution targets in 3D ($\Delta p$) and the horizontal plane ($\Delta p_{xy}$).
The approach is suitable: targets were reached with an average horizontal error below \SI{0.3}{\meter}, even at the highest tested velocity of \SI{21}{\meter\per\second} (\SI{75.6}{\kilo\meter\per\hour}).

\begin{table*}[!htb] 
    \centering
    \normalsize
    {\renewcommand{\tabcolsep}{5.8pt} 
    \renewcommand{\arraystretch}{0.75}
    \caption{Real-world validation results} 
    \label{tbl:rw}
    \begin{tabular}{cccccc}
    \toprule 
    \Vmax [\si{\m\per\s}] & \Amax [\si{\m\per\s\squared}] & RMSE[\si{\m}] & $\Delta p$[\si{\m}] & $\Delta p_{xy}$[\si{\m}]\\
    \midrule
    9 & 15 & 0.605 & 0.298 & 0.251\\
    15 & 15 & 0.754 & 0.462 & 0.274\\
    21 & 15 & 0.671 & 0.503 & 0.277\\

    \bottomrule 
    \end{tabular} 
    } 
    \vspace{-1.0em}
\end{table*}


\section{Conclusion}\label{sec:conclusion}

We presented the DVOP, a generalization of the OP that captures second-order PMM dynamics under acceleration and gravity constraints, to apply the OP to quad-rotor UAVs.
To tackle this NP-hard problem, which couples combinatorial optimization with minimum-time trajectory planning, we proposed two solvers: an LNS metaheuristic for fast solutions on large instances, and an exact BnB based on a novel MILP relaxation of point-mass dynamics.
On KOP benchmarks, both solvers improve on the state of the art by up to \SI{37}{\percent}.
Evaluation over random instances shows solution quality increasing with the solution space for both, with the LNS--BnB gap below \SI{18}{\percent}.
A real-world validation showed a 3D RMS tracking error below \SI{0.5}{\meter} during agile flight at \SI{21}{\meter\per\second} with \SI{15}{\meter\per\square\second} acceleration.

\section*{Acknowledgment}
This work was supported by the Czech Science Foundation (GAČR) under research project no. 26-22606S and the CTU grant no. SGS26/077/OHK3/1T/13.

\bibliographystyle{elsarticle-harv}
\bibliography{main}

@string{iros = "{IEEE/RSJ} International Conference on Intelligent Robots and Systems (IROS)"}

@string{icra = "{IEEE} International Conference on Robotics and Automation (ICRA)"}

@string{icuas = "International Conference on Unmanned Aircraft Systems (ICUAS)"}

@article{bnb_original_paper,
 ISSN = {00129682, 14680262},
 author = {A. H. Land and A. G. Doig},
 journal = {Econometrica},
 number = {3},
 pages = {497--520},
 publisher = {[Wiley, Econometric Society]},
 title = {An Automatic Method of Solving Discrete Programming Problems},
 urldate = {2025-03-18},
 volume = {28},
 year = {1960},
 doi={10.2307/1910129}
}

@article{boussier2007exact,
  title={An exact algorithm for team orienteering problems},
  author={Boussier, Sylvain and Feillet, Dominique and Gendreau, Michel},
  journal={4or},
  volume={5},
  pages={211--230},
  year={2007},
  publisher={Springer},
  doi={10.1007/s10288-006-0009-1}
}

@article{meyer2022kinematic,
  author={Meyer, Fabian and Glock, Katharina},
  journal={IEEE Robotics and Automation Letters}, 
  title={Kinematic Orienteering Problem With Time-Optimal Trajectories for Multirotor UAVs}, 
  year={2022},
  volume={7},
  number={4},
  pages={11402-11409},
  doi={10.1109/LRA.2022.3194688}}

@article{VANSTEENWEGEN20111,
    title = {The orienteering problem: A survey},
    journal = {European Journal of Operational Research},
    volume = {209},
    number = {1},
    pages = {1-10},
    year = {2011},
    issn = {0377-2217},
    doi = {10.1016/j.ejor.2010.03.045},
    author = {Pieter Vansteenwegen and Wouter Souffriau and Dirk Van Oudheusden},
}

@article{pvenivcka2017dubins,
  title={Dubins orienteering problem},
  author={P{\v{e}}ni{\v{c}}ka, Robert and Faigl, Jan and V{\'a}{\v{n}}a, Petr and Saska, Martin},
  journal={IEEE Robotics and Automation Letters},
  volume={2},
  number={2},
  pages={1210--1217},
  year={2017},
  publisher={IEEE},
  doi={10.1109/LRA.2017.2666261}
}

@article{nekovavr2023multi,
  title={Multi-vehicle dynamic water surface monitoring},
  author={Nekov{\'a}{\v{r}}, Franti{\v{s}}ek and Faigl, Jan and Saska, Martin},
  journal={IEEE Robotics and Automation Letters},
  volume={8},
  number={10},
  pages={6323--6330},
  year={2023},
  publisher={IEEE},
  doi={10.1109/LRA.2023.3304533}
}

@article{ghotavadekar2024variable,
  title={Variable Time-Step MPC for Agile Multi-Rotor UAV Interception of Dynamic Targets},
  author={Ghotavadekar, Atharva and Nekov{\'a}{\v{r}}, Franti{\v{s}}ek and Saska, Martin and Faigl, Jan},
  journal={IEEE Robotics and Automation Letters},
  year={2024},
  publisher={IEEE},
  doi={10.1109/LRA.2024.3518096}
}

@misc{qin2024time,
  title={Time-Optimal Planning for Long-Range Quadrotor Flights: An Automatic Optimal Synthesis Approach}, 
  author={Chao Qin and Jingxiang Chen and Yifan Lin and Abhishek Goudar and Angela P. Schoellig and Hugh H. -T. Liu},
  year={2024},
  eprint={2407.17944},
  archivePrefix={arXiv},
  primaryClass={cs.RO},
}

@article{teissing2024real,
  author={Teissing, Krystof and Novosad, Matej and Penicka, Robert and Saska, Martin},
  journal={IEEE Robotics and Automation Letters}, 
  title={Real-Time Planning of Minimum-Time Trajectories for Agile UAV Flight}, 
  year={2024},
  volume={9},
  number={11},
  pages={10351-10358},
  doi={10.1109/LRA.2024.3471388}}

@article{hehn2012performance,
  title={Performance benchmarking of quadrotor systems using time-optimal control},
  author={Hehn, Markus and Ritz, Robin and D’Andrea, Raffaello},
  journal={Autonomous Robots},
  volume={33},
  pages={69--88},
  year={2012},
  publisher={Springer},
  doi={10.1007/s10514-012-9282-3}
}

@Inbook{pisinger2019lns,
    publisher={Springer},
    series={International Series in Operations Research \& Management Science},
    booktitle={Handbook of Metaheuristics},
    author={David Pisinger and Stefan Ropke},
    title={Large Neighborhood Search},
    year={2019},
    month={June},
    pages={99-127},
    address={Cham},
    doi={10.1007/978-3-319-91086-4_4}
}

@article{pvenivcka2019variable,
  title={Variable neighborhood search for the set orienteering problem and its application to other orienteering problem variants},
  author={P{\v{e}}ni{\v{c}}ka, Robert and Faigl, Jan and Saska, Martin},
  journal={European Journal of Operational Research},
  volume={276},
  number={3},
  pages={816--825},
  year={2019},
  publisher={Elsevier},
  doi = {10.1016/j.ejor.2019.01.047},
}

@article{golden1987orienteering,
  title={The orienteering problem},
  author={Golden, Bruce L and Levy, Larry and Vohra, Rakesh},
  journal={Naval Research Logistics (NRL)},
  volume={34},
  number={3},
  pages={307--318},
  year={1987},
  publisher={Wiley Online Library},
  doi={10.1002/1520-6750(198706)34:3<307::AID-NAV3220340302>3.0.CO;2-D}
}

@INPROCEEDINGS{faigl2019bezierop,
  author={Faigl, Jan and Váňa, Petr and Pěnička, Robert},
  booktitle={2019 International Conference on Robotics and Automation (ICRA)}, 
  title={Multi-Vehicle Close Enough Orienteering Problem with Bézier Curves for Multi-Rotor Aerial Vehicles}, 
  year={2019},
  volume={},
  number={},
  pages={3039-3044},
  doi={10.1109/ICRA.2019.8794339}
}

@INPROCEEDINGS{faigl2017dubinsSOM,
  author={Faigl, Jan},
  booktitle={2017 12th International Workshop on Self-Organizing Maps and Learning Vector Quantization, Clustering and Data Visualization (WSOM)}, 
  title={Self-organizing map for orienteering problem with dubins vehicle}, 
  year={2017},
  volume={},
  number={},
  pages={1-8},
  doi={10.1109/WSOM.2017.8020017}
}

@INPROCEEDINGS{faria2024vsdop,
  author={Vinícius Faria, L. C. and Caio Ribeiro, C. G. and Macharet, Douglas G.},
  booktitle={2024 Latin American Robotics Symposium (LARS)}, 
  title={Variable-Speed Dubins Orienteering Problem}, 
  year={2024},
  volume={},
  number={},
  pages={1-6},
  doi={10.1109/LARS64411.2024.10786481}
}

@book{kohonen2001tsp_som,
  title        = {Self-Organizing Maps},
  author       = {Teuvo Kohonen},
  edition      = {3rd},
  year         = {2001},
  publisher    = {Springer Berlin, Heidelberg},
  doi={10.1007/978-3-642-56927-2}
}

@inproceedings{Meyer2023topuav,
  title={Top-uav: Open-source time-optimal trajectory planner for point-masses under acceleration and velocity constraints},
  author={Meyer, Fabian and Glock, Katharina and Sayah, David},
  booktitle={2023 IEEE/RSJ International Conference on Intelligent Robots and Systems (IROS)},
  pages={2838--2845},
  year={2023},
  organization={IEEE},
  doi={10.1109/IROS55552.2023.10342270}
}

@article{dubins1957dubins,
 author = {L. E. Dubins},
 journal = {American Journal of Mathematics},
 number = {3},
 pages = {497--516},
 publisher = {Johns Hopkins University Press},
 title = {On Curves of Minimal Length with a Constraint on Average Curvature, and with Prescribed Initial and Terminal Positions and Tangents},
 volume = {79},
 year = {1957},
 doi={10.2307/2372560}
}

@misc{hsl_collection,
  title={A collection of Fortran codes for large-scale scientific computation},
  author={HSL},
  year = {2022},
  url = {http://www.hsl.rl.ac.uk/}
}

@article{ipopt,
   title = {On the implementation of an interior-point filter line-search algorithm for large-scale nonlinear programming},
   author = {Wächter, Andreas and Biegler, Lorenz T.},
   journal = {Mathematical programming},
   volume = {106},
   number = {1},
   pages = {25--57},
   year = {2006},
   doi={10.1007/s10107-004-0559-y}
}

@article{Foehn2022AlphaPilot,
  author = {Philipp Foehn and Dario Brescianini and Elia Kaufmann and Titus Cieslewski and Mathias Gehrig and Manasi Muglikar and Davide Scaramuzza},
  journal = {Autonomous Robots},
  title = {AlphaPilot: autonomous drone racing},
  year = {2022},
  pages = {307-320},
  volume = {46},
  number = {1},
  ISSN = {0929-5593},
  doi={10.1007/s10514-021-10011-y}
}

@article{Romero2022PMMReplanningMPCC,
  author = {Angel Romero and Robert Penicka and Davide Scaramuzza},
  journal = {IEEE Robotics and Automation Letters},
  title = {Time-Optimal Online Replanning for Agile Quadrotor Flight},
  year = {2022},
  pages = {7730-7737},
  volume = {7},
  number = {3},
  ISSN = {2377-3766},
  doi={10.1109/LRA.2022.3185772}
}

@article{morandi2024orienteering,
  title={The orienteering problem with drones},
  author={Morandi, Nicola and Leus, Roel and Yaman, Hande},
  journal={Transportation Science},
  volume={58},
  number={1},
  pages={240--256},
  year={2024},
  publisher={INFORMS},
  doi={10.1287/trsc.2023.0003}
}

@article{dasdemir2022uav,
  title={UAV routing for reconnaissance mission: A multi-objective orienteering problem with time-dependent prizes and multiple connections},
  author={Dasdemir, Erdi and Batta, Rajan and K{\"o}ksalan, Murat and {\"O}zt{\"u}rk, Diclehan Tezcaner},
  journal={Computers \& Operations Research},
  volume={145},
  pages={105882},
  year={2022},
  publisher={Elsevier},
  doi={10.1016/j.cor.2022.105882}
}

@article{flood1956traveling,
  title={The traveling-salesman problem},
  author={Flood, Merrill M},
  journal={Operations research},
  volume={4},
  number={1},
  pages={61--75},
  year={1956},
  publisher={INFORMS},
  doi={10.1287/opre.4.1.61}
}

@article{kolesar1967branch,
  title={A branch and bound algorithm for the knapsack problem},
  author={Kolesar, Peter J},
  journal={Management science},
  volume={13},
  number={9},
  pages={723--735},
  year={1967},
  publisher={INFORMS},
  doi={10.1287/mnsc.13.9.723}
}

@article{sundar2022branch,
  title={A branch-and-price algorithm for a team orienteering problem with fixed-wing drones},
  author={Sundar, Kaarthik and Sanjeevi, Sujeevraja and Montez, Christopher},
  journal={EURO Journal on Transportation and Logistics},
  volume={11},
  pages={100070},
  year={2022},
  publisher={Elsevier},
  doi={10.1016/j.ejtl.2021.100070}
}

@article{psaraftis1988dynamic,
  title={Dynamic vehicle routing problems},
  author={Psaraftis, Harilaos N},
  journal={Vehicle routing: Methods and studies},
  volume={16},
  pages={223--248},
  year={1988},
  publisher={North Holland, Amsterdam, The Netherlands}
}

@article{peyman2024sim,
  title={A sim-learnheuristic for the team orienteering problem: applications to unmanned aerial vehicles},
  author={Peyman, Mohammad and Martin, Xabier A and Panadero, Javier and Juan, Angel A},
  journal={Algorithms},
  volume={17},
  number={5},
  pages={200},
  year={2024},
  publisher={MDPI},
  doi={10.3390/a17050200}
}

@article{gunawan2016orienteering,
  title={Orienteering problem: A survey of recent variants, solution approaches and applications},
  author={Gunawan, Aldy and Lau, Hoong Chuin and Vansteenwegen, Pieter},
  journal={European journal of operational research},
  volume={255},
  number={2},
  pages={315--332},
  year={2016},
  publisher={Elsevier},
  doi={10.1016/j.ejor.2016.04.059}
}

@inproceedings{Souriau2008AGR,
  title={A greedy randomised adaptive search procedure for the team orienteering problem},
  author={Souffriau, Wouter and Vansteenwegen, Pieter and Vanden Berghe, Greet and Van Oudheusden, Dirk},
  booktitle={EU/MEeting 2008 on metaheuristics for logistics and vehicle routing},
  volume={2008},
  pages={23--24},
  year={2008}
}

@InProceedings{zahradka2019dtopn,
    author="Zahr{\'a}dka, David
    and P{\v{e}}ni{\v{c}}ka, Robert
    and Saska, Martin",
    title="Route Planning for Teams of Unmanned Aerial Vehicles Using Dubins Vehicle Model with Budget Constraint",
    booktitle="Modelling and Simulation for Autonomous Systems",
    year="2019",
    publisher="Springer International Publishing",
    address="Cham",
    pages="365--389",
    isbn="978-3-030-14984-0",
    doi="10.1007/978-3-030-14984-0_27"
}

@INPROCEEDINGS{penicka2017dopn,
  author={Pěnička, Robert and Faigl, Jan and Váňa, Petr and Saska, Martin},
  booktitle={2017 International Conference on Unmanned Aircraft Systems (ICUAS)}, 
  title={Dubins orienteering problem with neighborhoods}, 
  year={2017},
  volume={},
  number={},
  pages={1555-1562},
  doi={10.1109/ICUAS.2017.7991350}}

@INPROCEEDINGS{faigl2017dceop,
  author={Faigl, Jan and Pěnička, Robert},
  booktitle={2017 IEEE/RSJ International Conference on Intelligent Robots and Systems (IROS)}, 
  title={On close enough orienteering problem with Dubins vehicle}, 
  year={2017},
  volume={},
  number={},
  pages={5646-5652},
  doi={10.1109/IROS.2017.8206453}}

@INPROCEEDINGS{vana2017dapop,
  author={Váňa, Petr and Faigl, Jan and Sláma, Jakub and Pěnička, Robert},
  booktitle={2017 European Conference on Mobile Robots (ECMR)}, 
  title={Data collection planning with Dubins airplane model and limited travel budget}, 
  year={2017},
  volume={},
  number={},
  pages={1-6},
  doi={10.1109/ECMR.2017.8098715}}

@INPROCEEDINGS{faigl2016dopnsom,
  author={Faigl, Jan and Pěnička, Robert and Best, Graeme},
  booktitle={2016 IEEE International Conference on Systems, Man, and Cybernetics (SMC)}, 
  title={Self-organizing map-based solution for the Orienteering problem with neighborhoods}, 
  year={2016},
  volume={},
  number={},
  pages={001315-001321},
  doi={10.1109/SMC.2016.7844421}}

@misc{gurobi,
  author = {{Gurobi Optimization, LLC}},
  title = {{Gurobi Optimizer Reference Manual}},
  year = 2026,
  url = "https://www.gurobi.com"
}

@article{mor2022vehicle,
  title={Vehicle routing problems over time: a survey},
  author={Mor, Andrea and Speranza, Maria Grazia},
  journal={Annals of Operations Research},
  volume={314},
  number={1},
  pages={255--275},
  year={2022},
  publisher={Springer},
  doi={10.1007/s10479-021-04488-0}
}

@article{mladenovic1997variable,
  title={Variable neighborhood search},
  author={Mladenovi{\'c}, Nenad and Hansen, Pierre},
  journal={Computers \& operations research},
  volume={24},
  number={11},
  pages={1097--1100},
  year={1997},
  publisher={Elsevier}
}

@inproceedings{shaw1998using,
  title={Using constraint programming and local search methods to solve vehicle routing problems},
  author={Shaw, Paul},
  booktitle={International conference on principles and practice of constraint programming},
  pages={417--431},
  year={1998},
  organization={Springer},
  doi={10.1007/3-540-49481-2_30}
}

@article{tsiligirides1984dataset,
     author = {T. Tsiligirides},
     journal = {The Journal of the Operational Research Society},
     number = {9},
     pages = {797--809},
     publisher = {Palgrave Macmillan Journals},
     title = {Heuristic Methods Applied to Orienteering},
     urldate = {2026-06-03},
     volume = {35},
     year = {1984},
     doi = {10.2307/2582629}
}

@article{chao1996dataset,
    title = {A fast and effective heuristic for the orienteering problem},
    journal = {European Journal of Operational Research},
    volume = {88},
    number = {3},
    pages = {475-489},
    year = {1996},
    issn = {0377-2217},
    doi = {10.1016/0377-2217(95)00035-6},
    author = {I-Ming Chao and Bruce L. Golden and Edward A. Wasil},
}

@article{baca2021mrs,
	author = "Baca, Tomas and Petrlik, Matej and Vrba, Matous and Spurny, Vojtech and Penicka, Robert and Hert, Daniel and Saska, Martin",
	doi = "10.1007/s10846-021-01383-5",
	issue = 1,
	journal = "Journal of Intelligent {\&} Robotic Systems",
	month = "May",
	number = 26,
	pages = "1--28",
	publisher = "Springer",
	title = "{The MRS UAV System: Pushing the Frontiers of Reproducible Research, Real-world Deployment, and Education with Autonomous Unmanned Aerial Vehicles}",
	type = "article",
	volume = 102,
	year = 2021
}

@article{hert2023mrs,
	author = "{Hert}, D. and {Baca}, T. and {Petracek}, P. and {Kratky}, V. and {Penicka}, R. and {Spurny}, V. and {Petrlik}, M. and {Vrba}, M. and {Zaitlik}, D. and {Stoudek}, P. and {Walter}, V. and {Stepan}, P. and {Horyna}, J. and {Pritzl}, V. and {Sramek}, M. and {Ahmad}, A. and {Silano}, G. and {Bonilla Licea}, D. and {Stibinger}, P. and {Nascimento}, T. and {Saska}, M.",
	doi = "10.1007/s10846-023-01879-2",
	issue = 64,
	journal = "Journal of Intelligent {\&} Robotic Systems",
	month = "July",
	pages = "1--34",
	publisher = "Springer",
	title = "{MRS Drone: A Modular Platform for Real-World Deployment of Aerial Multi-Robot Systems}",
	volume = 108,
	year = 2023
}

\end{document}